\documentclass[sigconf,preprint]{acmart}
\AtBeginDocument{%
  }

\setcopyright{acmlicensed}
\copyrightyear{2018}
\acmYear{2018}
\acmDOI{XXXXXXX.XXXXXXX}
\acmConference[Conference acronym 'XX]{Make sure to enter the correct
  conference title from your rights confirmation email}{June 03--05,
  2018}{Woodstock, NY}
\acmISBN{978-1-4503-XXXX-X/2018/06}
\usepackage[table]{xcolor}
\usepackage{longtable}
\usepackage{tabularx}
\usepackage{multirow}
\usepackage{fontawesome5}
\usepackage{pifont}
\usepackage{booktabs} %
\usepackage{adjustbox}
\usepackage{graphicx}
\usepackage{xcolor}
\usepackage{tcolorbox}
\usepackage{makecell}
\usepackage{subcaption}

\begin{document}

\title{Query as Anchor: Scenario-Adaptive User Representation via Large Language Model}

% Shared contribution marker for first two authors
\author{Jiahao Yuan$^{1,2,*}$, Yike Xu$^{1,*}$, Jinyong Wen$^1$, Baokun Wang$^{1,\dag}$, Ziyi Gao$^1$, Xiaotong Lin$^1$}
\author{Yun Liu$^1$, Xing Fu$^1$, Yu Cheng$^1$, Yongchao Liu$^1$, Weiqiang Wang$^1$, Zhongle Xie$^3$}

\thanks{*Both authors contributed equally to this research. This work was completed during Jiahao’s research internship at Ant Group.}
\thanks{$\dagger$Corresponding author}

\affiliation{
  \institution{\textsuperscript{1}Ant Group, \textsuperscript{2}East China Normal University, \textsuperscript{3}Zhejiang University}
  \country{China}
}
\authorsaddresses{
  \noindent $^*$ Both authors contributed equally to this research. \\
  \noindent $^\dag$ Corresponding author. \\
  \noindent \textsuperscript{1}Ant Group, \textsuperscript{2}Zhejiang University.
}

\renewcommand{\shortauthors}{Yuan et al.}

\begin{abstract}
Industrial-scale user representation learning requires balancing robust universality with acute task-sensitivity. However, existing \\ paradigms primarily yield static, task-agnostic embeddings that struggle to reconcile the divergent requirements of downstream scenarios within unified vector spaces. Furthermore, heterogeneous multi-source data introduces inherent noise and modality conflicts, degrading representation. We propose Query-as-Anchor, a framework shifting user modeling from static encoding to dynamic, query-aware synthesis. To empower Large Language Models (LLMs) with deep user understanding, we first construct UserU, an industrial-scale pre-training dataset that aligns multi-modal behavioral sequences with user understanding semantics, and our Q-Anchor Embedding architecture integrates hierarchical coarse-to-fine encoders into dual-tower LLMs via joint contrastive-autoregressive optimization for query-aware user representation. To bridge the gap between general pre-training and specialized business logic, we further introduce Cluster-based Soft Prompt Tuning to enforce discriminative latent structures, effectively aligning model attention with scenario-specific modalities. For deployment, anchoring queries at sequence termini enables KV-cache-accelerated inference with negligible incremental latency. Evaluations on 10 Alipay industrial benchmarks show consistent SOTA performance, strong scalability, and efficient deployment. Large-scale online A/B testing in Alipay's production system across two real-world scenarios further validates its practical effectiveness. Our code is prepared for public release and will be available at:
\url{https://github.com/JhCircle/Q-Anchor}.

\end{abstract}





\received{20 February 2007}
\received[revised]{12 March 2009}
\received[accepted]{5 June 2009}

\maketitle

\section{Introduction}
User representation learning underpins modern industrial intelligence systems by enabling personalized, data-driven decision making across applications such as recommendation~\cite{chai2022user}, digital marketing~\cite{zhang2024scaling}, and risk management~\cite{dou2025transferable}. In practice, users frequently engage in multiple business scenarios, each characterized by distinct behavioral patterns and decision objectives, which calls for representations that are both transferable across tasks and adaptable to scenario-specific contexts. Accordingly, existing systems typically compress large-scale, heterogeneous user signals, including textual profiles, interaction histories, and structured attributes, into compact embeddings to support downstream modeling and inference.

Most existing user embedding methods learn static, task-specific representations, either trained from scratch~\cite{fu2023robust} or based on pretrained language model (PLM) encoders optimized with contrastive objectives~\cite{sun2019bert4rec, oord2018representation}. While effective at capturing historical behavioral patterns, these approaches are inherently retrospective and lack the flexibility to adapt to diverse downstream decision scenarios. As a result, industrial systems often rely on multiple task-specific user models for different applications, increasing system complexity, deployment overhead, and long-term maintenance cost, while still struggling with cross-domain generalization~\cite{li2023one}.

Recent research has explored large language models (LLMs) as user encoders, enhancing sequential and semantic understanding through fine-tuning on behavior sequences~\cite{sun2019bert4rec, ning2025user} or by enriching user and item semantics~\cite{hu2024enhancing}. However, real-world user behaviors are typically sparse, symbolic, and highly heterogeneous, which diverges significantly from the dense, language-centric data used in LLM pretraining~\cite{zhang2025qwen3,babakhin2025llama,hu2025kalm}. This modality and semantic gap limits the ability of LLMs to generate adaptive and anticipatory user representations, especially in specialized domains where behavior distributions differ substantially from pretraining corpora. While instruction-aware embeddings~\cite{zhang2025qwen3} offer limited task conditioning, they remain insufficient for capturing the complex, dynamic, and noisy nature of industrial user data.

Overall, existing user representation learning methods face three key challenges: 
\textbf{(1) Limited scenario adaptability and cross-task generalization:} Static embeddings cannot flexibly support diverse downstream tasks such as credit assessment, marketing optimization, and risk management, each of which requires scenario-specific user understanding~\cite{tang2024we,zhao2024can}. 
\textbf{(2) Modality and semantic gap:} The symbolic sparsity and heterogeneous structure of behavioral logs are misaligned with language-centric pretraining, constraining LLM-based user representations from effectively generalizing across domains. 
\textbf{(3) Heterogeneous data integration at scale:} Industrial user data varies significantly in relevance across scenarios, requiring mechanisms to selectively attend to scenario-relevant signals, suppress noise, and compress large-scale behaviors without introducing prohibitive inference overhead.

To address these challenges, we propose a unified user representation learning framework that integrates industrial-scale pretraining with efficient scenario adaptation. We introduce \textbf{Query-as-Anchor}, a query-conditioned mechanism that separates user behavior encoding from scenario-specific objectives. Our key contributions are summarized as follows:
\begin{itemize}
    \item We construct \textbf{\textsc{UserU}}, an industrial-scale pretraining dataset that aligns heterogeneous user behaviors with user understanding semantics via future behavior prediction and QA-based supervision, providing strong behavioral and semantic priors for user embedding learning.
    \item We propose \textbf{\textsc{Query-as-Anchor}}, a query-conditioned framework that generates scenario-adaptive user embeddings by re-anchoring the same behavioral profile under different downstream decision contexts, enabling reuse across multiple business scenarios.
    \item We introduce \textbf{soft-prompt tuning} and \textbf{KV-cache--aware acceleration}, enabling efficient scenario specialization and low-latency multi-scenario inference without retraining the backbone model.
    \item Extensive offline experiments and large-scale online A/B testing on Alipay demonstrate consistent performance gains across user engagement, risk control, and marketing scenarios, while significantly reducing system complexity and deployment overhead.
\end{itemize}

\section{Related Work}
\paragraph{LLM for User Embedding.}
Large language models (LLMs) have demonstrated significant potential in user representation learning by integrating diverse data types, such as textual profiles, behavioral sequences, and structured attributes, into unified embeddings. Early works like BERT4Rec~\cite{sun2019bert4rec} and FOUND~\cite{dou2025transferable} fine-tuned LMs \cite{peng2024answer} on behavioral data to capture temporal dependencies via masked prediction and contrastive objectives \cite{gao2021simcse}, but they rely on static user embeddings \cite{lin2022dual}, which limits adaptability to dynamic or evolving contexts. More recent instruction-aware models, such as Qwen3-embedding~\cite{zhang2025qwen3} and KaLM-Embedding \cite{hu2025kalm}, have extended LLMs for task-specific embeddings \cite{behnamghaderllm2vec,he2025llm2rec}. However, these models still face challenges when transitioning from general-purpose pretraining on large text corpora to sparse, symbolic, and highly contextual user behavior data, creating a gap in both structure and semantics. In contrast, our Query as Anchor framework bridges this gap by introducing dynamic, query-aware embeddings that adapt to the evolving nature of user behaviors, improving adaptability and contextual sensitivity across diverse tasks and domains.

\begin{figure}[t]
\centering
\includegraphics[width=\linewidth]{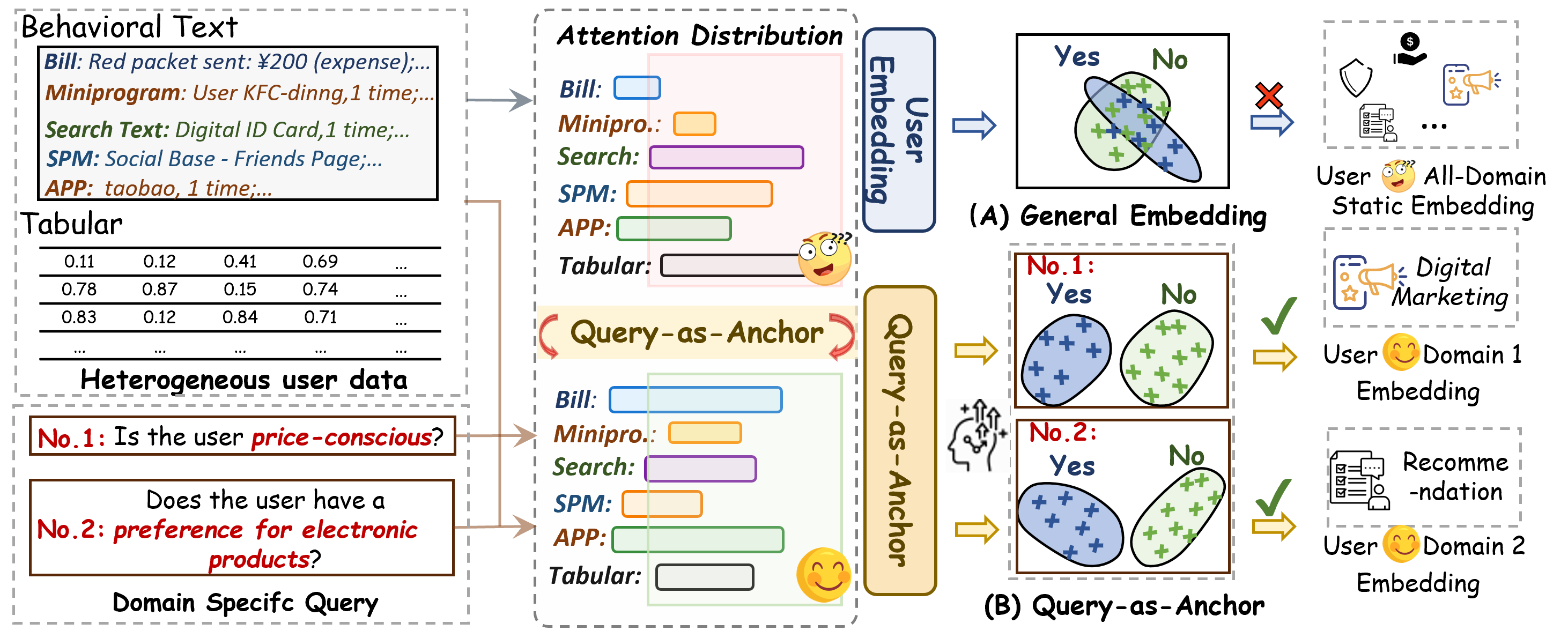} 
\caption{Comparison between (A) General User Embedding \cite{dou2025transferable} and (B) our Query-as-anchor. (A) learns transferable user representations across domains but generates fixed embeddings regardless of downstream context. (B) extends (A) with query-as-anchor modulation, enabling a single model to produce adaptive, domain-specific embeddings via natural language instructions.}
\label{Fig.frame}
\end{figure}

\paragraph{Synthetic Data for User Embedding.}

Despite advancements in user representation learning, a significant gap remains due to the lack of a large-scale, comprehensive dataset specifically designed for user embedding pretraining. This limitation has spurred interest in generating synthetic data to supplement the training process. Early methods primarily focused on heuristic data augmentation and pseudo-labeling approaches~\cite{nogueira2019passage,yuan2025kardia}, which aimed to simulate user behaviors and interactions. More recently, large language models (LLMs) have been employed to generate realistic user behavior, intent patterns, and interaction sequences \cite{dou2025transferable}, offering scalable and diverse data generation capabilities. However, these techniques still face challenges in capturing the full complexity of user behaviors. The absence of a dedicated pretraining dataset for user embeddings remains a critical obstacle. To overcome this, we introduce UseU, a large-scale dataset designed specifically for user embedding pretraining. By incorporating rule-based future behavior prediction and QA-understanding tasks, UseU enables more effective, context-aware learning of user representations. 
\section{Methodology}

\begin{figure*}[t]
\centering
\includegraphics[width=\textwidth]{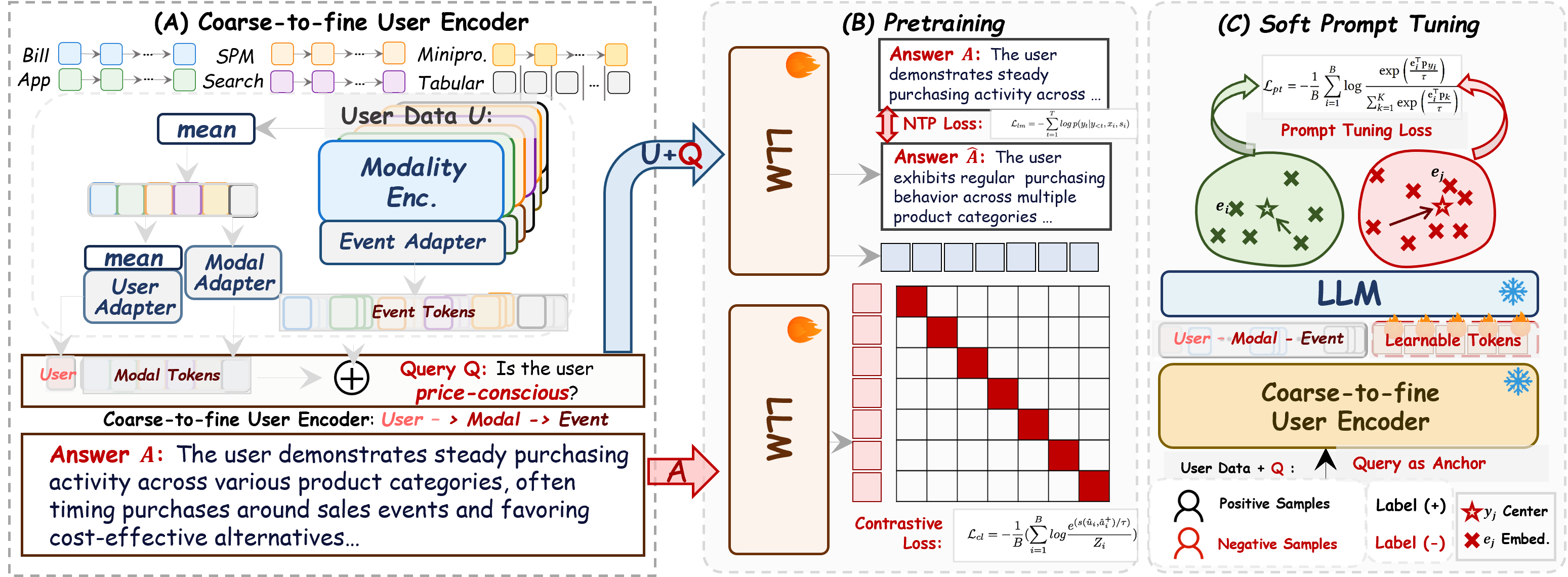} 
\caption{Overview of Query-as-Anchor Framework for our Q-Anchor Embedding.}
\label{fig:method}
\end{figure*}

We propose \textsc{Q-Anchor Embedding}, a large-scale pretraining framework for user understanding within the Alipay ecosystem. Here, $\mathbf{u}_i=\{Bill_i, Mini_i, Spm_i, App_i, Search_i, Tabular_i\}\in\mathcal{U}$ denotes user $i$'s multi-modal interaction profile over the past 90 days, where $Bill_i$ are PayBill transactions, $Mini_i$ are Mini Program interactions, $Spm_i$ are SPM navigation paths, $App_i$ are app-level list, $Search_i$ are on-platform search queries, and $Tabular_i\in\mathbb{R}^{1\times F}$ contains $F$ structured features. This unified profile constitutes the input to our UserU pretraining dataset (§~\ref{subsec:dataset}) and the Query-as-Anchor framework (§~\ref{subsec:anchor}), which aligns heterogeneous behaviors via query-driven anchors. To support efficient transfer, we further apply soft prompt tuning to adapt anchor embeddings to diverse downstream tasks with minimal parameter updates.

\subsection{UserU Pretraining Dataset}
\label{subsec:dataset}
To enhance user embedding performance across diverse tasks, we introduce the User Understanding (UserU) Pretraining Dataset, which integrates dynamic, context-aware user behavior with task adaptability for real-world applications. Extending future prediction pretraining \cite{dou2025transferable}, we augment it with synthesized query-answer pairs to capture deeper user understanding. UserU combines two key components: a behavior prediction dataset for future user actions $D_{future}$ and a LLM-generated UserQA dataset $D_{uqa}$ for comprehensive user representation via query-answer synthesis. Both data examples are detailed in Appendix~\ref{app:dataexample}.

\paragraph{Behavior-based Interaction Dataset $ \mathcal{D}_{future} $.} We construct a behavior- grounded supervision signal by pairing each user's historical profile with a \emph{future-action summary} in a unified query--answer format (Appendix~\ref{app:dataexample}). Concretely, for each user $i$, we build a three-month behavior profile $\mathbf{u}_i$ from multi-modal logs. We then derive a target $a_i^{\mathrm{future}}$ by aggregating subsequent interactions into temporal bins and action categories, and selecting the frequency- and diversity-aware actions as the representative subset. Using a fixed template query $q^{\mathrm{future}}$ (e.g., ``What are the user's most likely actions in the next period?''), we form training pairs
$\mathcal{D}_{future}=\{(\mathbf{u}_i \oplus q^{\mathrm{future}},\, a_i^{\mathrm{future}})\}_{i=1}^{N}$, where $\oplus$ denotes contextual concatenation. This dataset encourages the embedding to capture temporal regularities and encode predictive signals for near-future behaviors.
\paragraph{Synthetic Query-Answer Dataset $ \mathcal{D}_{uqa} $.} To address the scarcity of user understanding training data and ensure better decoupling between pretraining data and downstream tasks for improved generalization, we propose a self-reflect synthetic data generation approach to construct a general-purpose user-query-answer alignment dataset. Inspired by the cold-start strategy in~\cite{chen2025little}, we first initialize a seed pool $\mathcal{P}$ comprising 72 life-related user-understanding topics (e.g., financial planning, health management) via Qwen-Max. Given a user profile $\mathbf{u}_i$, we prompt an LLM $\mathcal{M}$ to retrieve the top-10 most relevant topics from $\mathcal{P}$, and instantiate each topic into a naturalistic query $q_i$ grounded in $\mathbf{u}_i$. Conditioned on $(\mathbf{u}_i, q_i)$, the LLM then generates an answer $a_i$. To improve faithfulness and contextual validity, we apply a post-generation reflection step in which $\mathcal{M}$ re-checks the draft answer against $\mathbf{u}_i$ and revises unsupported or inconsistent statements. The resulting dataset is
$\mathcal{D}_{uqa}=\{(\mathbf{u}_i \oplus q_i, a_i)\}_{i=1}^{M}$,
where $\oplus$ denotes contextual concatenation.

\subsection{Query as Anchor}
\label{subsec:anchor}

\subsubsection{Hierarchical Coarse-to-fine User Encoder.}

To effectively reconcile the inherent sparsity of multi-source behavioral signals with the dense semantic requirements of Large Language Models (LLMs), we propose a hierarchical encoding architecture that distills raw interactions into a multi-granularity representation space. Specifically, for each modality $m \in \mathcal{M}$ (where $\mathcal{M} =\{Bill, Mini, SPM, App, $  \\ $Search,
Tabular\}$), raw event sequences are first projected into initial embeddings $\{\mathbf{h}_{m,t}\}$ via encoder and subsequently refined by modality-specific event adapters:
\begin{equation}
    \mathbf{z}_{m,t}^{(evt)} = \text{MLP}_m^{(evt)}\left(\text{LayerNorm}(\mathbf{h}_{m,t})\right),
\end{equation}
thereby preserving fine-grained atomic features of individual actions. To capture intra-modality trends while mitigating idiosyncratic noise, these event-level embeddings are aggregated through mean-pooling into a summary vector $\bar{\mathbf{z}}_m^{(evt)}$, which is further transformed by a shared modal adapter to yield a unified modality embedding $\mathbf{z}_m^{(mdl)} = \text{MLP}^{(mdl)}(\text{LayerNorm}(\bar{\mathbf{z}}_m^{(evt)}))$. At the apex of this hierarchy, a global user-level representation $\mathbf{z}^{(usr)}$ is derived by consolidating all modality-specific vectors through a dedicated user adapter, capturing the holistic behavioral profile. The final comprehensive input tokens $\mathbf{e}_i$ is constructed via the structured concatenation of representations across all three levels:
\begin{equation}
\label{eq:tokens}
\mathbf{e}_i
= \Big[\, \mathbf{z}^{(\mathrm{usr})} \ ;\ 
\{\mathbf{z}^{(\mathrm{mdl})}_m\}_{m\in\mathcal{M}} \ ;\ 
\{ \bar{\mathbf{z}}^{(\mathrm{evt})}_{m} \}_{m\in\mathcal{M}}^{K_m} \,\Big],
\end{equation}

where $[\cdot;\cdot]$ denotes the concatenation operation and $K_m$ is the number of event tokens retained for modality $m$. This hierarchical design lets the LLM attend to either fine-grained events or high-level behavior summaries conditioned on the query, while remaining compatible with its native embedding space.

\begin{figure}[t]
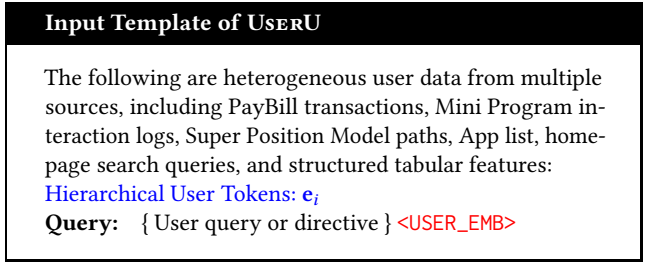

\centering
\begin{tcolorbox}[
    width=\linewidth,
    colback=white,
    colframe=black,
    title=Input Template of \textsc{UserU},
    fonttitle=\bfseries\normalsize,
    sharp corners,
    boxrule=1pt,
    top=6pt,
    bottom=6pt
]
% \small
The following are heterogeneous user data from multiple sources, including PayBill transactions, Mini Program interaction logs, Super Position Model paths, App list, homepage search queries, and structured tabular features:

\textcolor{blue}{ Hierarchical User Tokens: $\mathbf{e}_i$}

\noindent
\textbf{Query:}\quad \{ \textnormal{User query or directive} \} \textcolor{red}{\texttt{<USER\_EMB>}}
\end{tcolorbox}
\caption{
Input Template of \textsc{UserU}. The \textcolor{blue}{Hierarchical User Tokens: $\mathbf{e}_i$} injects the precomputed hierarchical embedding $\mathbf{e}_i$ (Eq.~\ref{eq:tokens}). An optional instruction is followed by the special \textcolor{red}{\texttt{<USER\_EMB>}} token, which signals the model to extract a unified user embedding (Sec.~\ref{subsubsec:pretrain}).
}
\label{fig:input_format}
\end{figure}

\subsubsection{Q-Anchor Pretraining Architecture.}
\label{subsubsec:pretrain}
\paragraph{Query-as-Anchor Dual-Tower Architecture.} Building upon the hierarchical user representations, we propose a dual-tower training architecture that operationalizes the \textit{Query-as-Anchor} paradigm by aligning multi-modal behavioral signals with semantic task directives as illustrated in Fig.~\ref{fig:method}. The primary \textit{Anchor Tower} ingests the hierarchical user tokens $\mathbf{e}_i$ and appends the natural language query $q_i$ as a trailing semantic anchor. By positioning the query at the sequence terminus, the LLM backbone acts as a query-aware aggregator that selectively distills intent-relevant features from the latent space of $\mathbf{e}_i$, ultimately projecting them into the anchored representation $\mathbf{u}_{i,q} = \text{LLM}_{anc}(\mathbf{e}_i, q_i)$. This structural decoupling is specifically engineered for industrial scalability; the computationally intensive hierarchical prefix $\mathbf{e}_i$ is computed once and stored via KV-cache mechanisms, allowing the model to generate diverse, scenario-adaptive embeddings for multiple queries with negligible incremental latency. Concurrently, an asymmetric \textit{Semantic Tower} projects the target answer $a_i$ into a dense vector $\mathbf{v}_{a_i} = \text{LLM}_{sem}(a_i)$, providing a high-fidelity linguistic target that serves as the ground-truth for user behavior synthesis and intent modeling. Specially, both towers share the same LLM parameters, which ensures that the behavioral features and semantic labels are mapped into a unified latent space.

\paragraph{Joint Contrastive--Generative Optimization.}
Our Query-as-Anchor framework is trained with a joint objective that combines discriminative contrastive alignment with generative grounding to produce user embeddings that are both distinctive and semantically rich. For each user $i$, the anchor tower generates a query-anchored embedding $\mathbf{u}_{i,q}$ from the hierarchical profile $\mathbf{e}_i$ and the task-specific query $q_i$, while the semantic tower encodes the answer $a_i$ into $\mathbf{v}_{a_i}$. Contrastive alignment is implemented via a query-conditioned InfoNCE loss:
\begin{small}
\begin{align}
\mathcal{L}_{cl} &= -\frac{1}{B} \sum_{i=1}^{B} 
\log \frac{\exp(\mathrm{sim}(\mathbf{u}_{i,q}, \mathbf{v}_{a_i}) / \tau)}{Z_i}, \\
Z_i &= \exp(\mathrm{sim}(\mathbf{u}_{i,q}, \mathbf{v}_{a_i}) / \tau)
+ \sum_{j \neq i} m_{ij} \exp(\mathrm{sim}(\mathbf{u}_{i,q}, \mathbf{v}_{a_j}) / \tau) \nonumber \\
&\quad + \sum_{j \neq i} m_{ij} \exp(\mathrm{sim}(\mathbf{u}_{i,q}, \mathbf{u}_{j,q}) / \tau)
+ \sum_{j \neq i} m_{ij} \exp(\mathrm{sim}(\mathbf{v}_{a_i}, \mathbf{v}_{a_j}) / \tau),
\end{align}
\end{small}

where $\mathrm{sim}(\cdot)$ is cosine similarity, $\tau$ is a temperature, $B$ is the batch size, and $m_{ij}$ is a margin-based mask that removes potential false negatives inspired by \cite{zhang2025qwen3}. Specifically, we discard a candidate negative $j$ for anchor $i$ if \emph{either} its user embedding or its answer embedding is too similar to $\mathbf{u}_{i,q}$:

\begin{small}
\begin{equation}
m_{ij}=
\begin{cases}
0, & \text{if }\ 
\exists\, x \in \{\mathbf{u}_{j,q},\,\mathbf{v}_{j}\}\ \text{s.t.}\
\mathrm{sim}(\mathbf{u}_{i,q}, x) >
\mathrm{sim}(\mathbf{u}_{i,q}, \mathbf{v}_{i}) + c_{\mathrm{margin}},\\
1, & \text{otherwise}.
\end{cases}
\end{equation}
\end{small}

To bridge the granularity gap between sentence-level alignment and token-level grounding \cite{li2025conan}, we propose a joint optimization objective that mitigates representation collapse while enhancing semantic density. While the contrastive loss $\mathcal{L}_{cl}$ enforces global discriminativeness by aligning positive pairs in the latent space, it often overlooks the fine-grained linguistic nuances essential for complex intent modeling. To counteract this, we introduce an auxiliary Next-Token Prediction (NTP) task that requires the anchor tower to autoregressively reconstruct the target answer $a_i$:
\begin{small}
\begin{equation}
\mathcal{L}_{ntp} = - \sum_{t=1}^{T} \log P(y_t \mid y_{<t}, e_i, q_i),
\end{equation}
\end{small}

where $T$ is the sequence length. The total objective is a weighted sum $\mathcal{L}_{total} = \mathcal{L}_{cl} + \mathcal{L}_{ntp}$, enabling the query to anchor the hierarchical user profile $\mathbf{e}_i$ into a compact, scenario-adaptive embedding that is both discriminative for downstream tasks and semantically rich for answer reconstruction.

\subsubsection{Soft Prompt Tuning.}
To further bridge the semantic gap between general-purpose user understanding and specialized downstream business logic, we introduce a cluster-based soft prompt tuning mechanism as a post-training adaption inspired by \cite{liu2022p,xiong2024dual}. As illustrated in Fig.~\ref{fig:method} (C), while the LLM backbone and the coarse-to-fine user encoder remain frozen to preserve the foundational multi-modal alignment, we introduce a set of learnable prompt tokens that function as differentiable task controllers. These tokens modulate the latent space of the hierarchical embeddings $\mathbf{u}_{i,q}$ to better align with downstream class-specific logic. We optimize these learnable tokens and a set of class prototypes $\{\mathbf{p}_k\}_{k=1}^K$ using a prototypical contrastive loss $\mathcal{L}_{pt}$, which pulls user embeddings toward their respective category centers while pushing them away from irrelevant clusters:
\begin{small}
\begin{equation}
\mathcal{L}_{pt} = -\frac{1}{B} \sum_{i=1}^{B} \log \frac{\exp\left(\frac{\mathbf{u}_{i,q}^\top \mathbf{p}_{y_i}}{\tau}\right)}{\sum_{k=1}^{K} \exp\left(\frac{\mathbf{u}_{i,q}^\top \mathbf{p}_k}{\tau}\right)},
\end{equation}
\end{small}

where $\mathbf{u}_{i,q}$ denotes the prompt-conditioned user embedding for sample $i$, $y_i$ is its ground-truth label, and $\mathbf{p}_k$ represents the learnable prototype (center) for class $k$. By maximizing the similarity between $\mathbf{u}_{i,q}$ and its corresponding class center $\mathbf{p}_{y_i}$ relative to all $K$ prototypes, the framework enforces a discriminative clustering structure in the latent space. This strategy enables the model to bridge the semantic gap between general pretraining and specialized business labels—such as high-risk vs. low-risk users—without the catastrophic forgetting associated with full parameter fine-tuning, thereby maintaining the structural integrity and deployment efficiency of the \textit{Query-as-Anchor} paradigm.

\begin{figure}[t]
\centering
\includegraphics[width=0.6\linewidth]{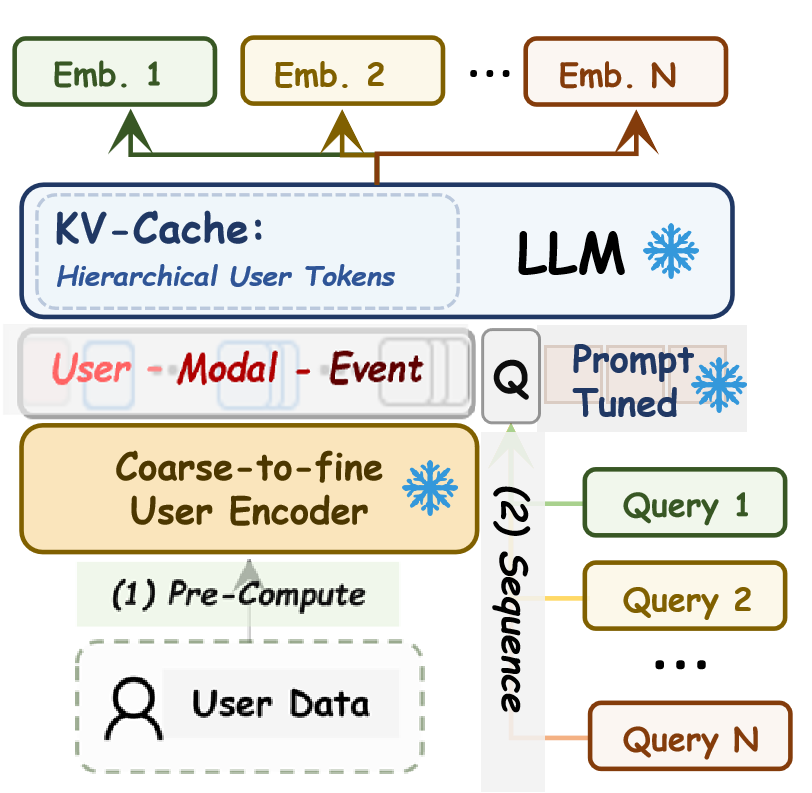} 
\caption{KV-Cache optimized Query-as-Anchor inference: pre-computed user prefixes enable sequence, low-latency re-anchoring across diverse tasks.}
\label{fig:deploy}
\end{figure}
\subsubsection{Query-as-Anchor: Accelerating Multi-Scenario Inference.}
To meet the demands of industrial deployment, \textsc{Q-Anchor} adopts a KV-cache optimization that decouples user encoding from task querying.
As shown in Fig.~\ref{fig:deploy}, for a given user $i$, the hierarchical profile $\mathbf{e}_i$ is encoded \emph{once} to produce a \textit{shared-context KV cache}. This cache is kept fixed as a persistent semantic prefix during inference. Given a set of downstream queries $\{q_1,\dots,q_n\}$, we process them sequentially while reusing the same shared-context cache.
For each query $q_j$, the model only computes the incremental hidden states for the short query tokens, resulting in an amortized complexity of $O(L_{q_j})$ per task (and $O(\sum_{j=1}^n L_{q_j})$ for all queries).
This design enables a single comprehensive user representation to be efficiently re-anchored to many business scenarios with negligible per-scenario incremental latency, supporting high-throughput embedding generation in the Alipay ecosystem.
Deployment details are provided in Appendix~\ref{app:deploy}.

\section{Experiments}

\paragraph{Models and Implementation.}
We adopt Qwen2.5-0.5B-Instruct~\cite{team2024qwen2} as a decoder-only backbone LLM for user representation learning. Heterogeneous user behaviors are encoded by modality-specific \texttt{gte-base}~\cite{li2023towards} encoders into dense embeddings, which are aggregated into unified user representations and fed into the backbone LLM for both training and inference. All baselines and ablations are trained under identical conditions: 50k fine-tuning steps with a global batch size of 2,048, using AdamW with an initial learning rate of $2 \times 10^{-4}$ and cosine decay. We employ LoRA~\cite{hulora} with rank 64 and $\alpha=32$ for pretraining inspired by \cite{zhang2025qwen3}, and user representations are fixed at 128 dimensions. Pretraining is conducted on 64 A100-80G GPUs with data parallelism, inference is performed on single A100-80G GPU for evaluation, and soft prompt tuning with 6 learnable tokens is carried out on single A100-80G GPU.  Pretraining scalability (data/model scale) and prompt tuning scalability (training steps/prompt tokens) are detailed in Appendix~\ref{subapp:per_scenario_base} and~\ref{subapp:per_scenario_pt}, respectively.

\paragraph{Baselines and Tasks.} We compare against two categories of baselines. \textbf{(1) General text embeddings} encode natural language descriptions of user behaviors, including Qwen2.5-0.5B-Instruct without user-specific fine-tuning and several top-ranked embedding models on MTEB including KaLM-Embedding-Gemma3-12B-2511 \cite{hu2025kalm}, llama-embed-nemotron-8b \cite{babakhin2025llama}, Qwen3-Embedding-8B \cite{zhang2025qwen3}. \textbf{(2) User representation models} include contrastive approaches such as MSDP~\cite{fu2023robust}, One4all \cite{shin2021one4all} and CPC~\cite{oord2018representation}, as well as LLM-based foundation models FOUND~\cite{dou2025transferable}. To study scalability, we also evaluate larger variants of our architecture via  under the same training protocol. We evaluate all methods on 10 real-world binary classification tasks from Alipay's production systems, grouped into three domains (Table~\ref{tab:task}). Detailed positive and negative label definitions for each task are provided in Appendix~\ref{app:dataset_details} (Table~\ref{tab:task_definitions}).
\begin{table}[!h]
\footnotesize
\centering
    \caption{Data information for user pretraining and test benchmarks, with number of tests per task.}
    \begin{tabular*}{\linewidth}{@{\extracolsep{\fill}} p{0.08\linewidth} p{0.2\linewidth} p{0.42\linewidth} p{0.15\linewidth} }
    \hline
         Dataset & Domain & Scenario & Number \\
         \hline
         $\mathcal{D}_{train}$ 
         & General~(\ref{subsec:dataset}) 
         & General 
         & $\approx$ $1.024 \times 10^8$ \\
         \hline
         \multirow{3}{*}{$\mathcal{D}_{test}$} 
         & \ding{182} User Engagement 
         & Interest Community Active User Identification (Active), Concert Click Prediction (Concert), User Log-in Prediction (Login), Ant Forest Engagement (Forest)
         & $\approx$ 50w per task \\
         & \ding{183} Risk 
         & Fraud Detection (Fraud), Money Laundering Detection (Money)
         & $\approx$ 50w per task \\
         & \ding{184} Marketing Se- nsitivity 
         & Takeout Interest (Takeout), Brand Sensitivity (Brand), Big Sale Sensitivity (Promo), Cost-Performance Sensitivity (Value)
         & $\approx$ 50w per task \\
         \hline
    \end{tabular*}

    \label{tab:task}
\end{table}

\paragraph{Evaluation Metrics.} Following \cite{dou2025transferable}, we assess representation via linear probing on 10 binary classification tasks from Alipay's user cognition system. We report AUC (Area Under the ROC Curve~\cite{bradley1997use}) for discriminative performance and KS (Kolmogorov-Smirnov~\cite{berger2014kolmogorov}) for critical decision-boundary separation. 

\section{Main Results}

\subsection{In-depth Performance Analysis}
\begin{table*}[!t]
\centering
\footnotesize
\caption{AUC performance on 10 key scenarios where our Q-Anchor (base version: pretrained) and (prompt-tuning version: post-trained) shows consistent improvement. KS results are detailed in Appendix~\ref{subapp:ks}.}
\label{tab:results}
\begin{tabular*}{\textwidth}{@{\extracolsep{\fill}} l *{11}{c} @{}}
\toprule
& \multicolumn{4}{c}{\textbf{User Engagement}} 
& \multicolumn{2}{c}{\textbf{Risk}} 
& \multicolumn{4}{c}{\textbf{Marketing Sensitivity}} 
& \textbf{Avg. AUC} \\
\cmidrule(lr){2-5} \cmidrule(lr){6-7} \cmidrule(lr){8-11}

\textbf{Method} 
& \textbf{Active} 
& \textbf{Concert} 
& \textbf{Login} 
& \textbf{Forest} 
& \textbf{Fraud} 
& \textbf{Money} 
& \textbf{Takeout} 
& \textbf{Brand} 
& \textbf{Promo} 
& \textbf{Value}
& \\
\midrule

\multicolumn{12}{@{}l@{}}{\textit{\textbf{General Embedding Models}}} \\
Qwen2.5-0.5B-Instruct
& 0.5269
& 0.5173
& 0.7219
& 0.7161
& 0.6969
& 0.6885
& 0.6361
& 0.5855
& 0.5483
& 0.7134
& 0.6351 \\

Qwen3-Embedding-0.6B     
& 0.5378
& 0.5226
& 0.7294
& 0.7287
& 0.7106
& 0.7123
& 0.6914
& 0.6076
& 0.5508
& 0.7172
& 0.6508
 \\
Llama-Embed-Nemotron-8B
& 0.5882 
& 0.5627
& 0.7735
& 0.8372
& 0.8632
& 0.8708
& 0.8176
& 0.7525
& 0.5918
& 0.8303
& 0.7488 \\

KaLM-Embed.-Gemma3-12B		
& 0.5597
& 0.5359
& 0.7609
& 0.7729
& 0.8229
& 0.8174
& 0.7812
& 0.6564
& 0.589
& 0.7609
& 0.7058 \\

\midrule
\multicolumn{12}{@{}l@{}}{\textit{\textbf{User Embedding Models}}} \\
MSDP \cite{fu2023robust} 
& 0.6415
& 0.5155
& 0.9504
& 0.9580
& 0.9152
& 0.8746
& 0.7814
& 0.6318
& 0.5850
& 0.8389
& 0.7692
\\
One4all \cite{shin2021one4all} 
& 0.6515
& 0.5568
& \textbf{0.9509}
& 0.9614
& 0.9203
& 0.8782
& 0.7609
& 0.6289
& 0.5761
& 0.8207
& 0.7706 \\
CPC~\cite{oord2018representation} 
& 0.6506
& 0.5314
& \underline{0.9506}
& 0.9608
& 0.9171
& 0.8736
& 0.7817
& 0.6235
& 0.6101
& 0.8259
& 0.7725
 \\
FOUND \cite{dou2025transferable}
& 0.6201
& 0.5527
& 0.8131
& 0.9573
& 0.9083
& 0.9235
& 0.8528
& 0.7294
& \underline{0.6202}
& 0.8544
& 0.7832 \\

\midrule
\multicolumn{12}{@{}l@{}}{\textit{\textbf{Q-Anchor Embedding (Ours)}}} \\
\textsc{Q-Anchor} (Base)    
& \underline{0.6568} & \underline{0.5739} & 0.8420 & \underline{0.9700} & \underline{0.9218} & \underline{0.9382} & \underline{0.8799} & \underline{0.7979} & 0.6189 & \underline{0.9049} & \underline{0.8104} \\
\textsc{Q-Anchor} (Prompt Tuned)      
& \textbf{0.6678} & \textbf{0.5844} & 0.8443 & \textbf{0.9716} & \textbf{0.9242} & \textbf{0.9439} & \textbf{0.8811} & \textbf{0.8535} & \textbf{0.6350} & \textbf{0.9194} & \textbf{0.8225} \\

\bottomrule
\end{tabular*}
\end{table*}

\begin{figure}[t]
\centering
\includegraphics[width=\linewidth]{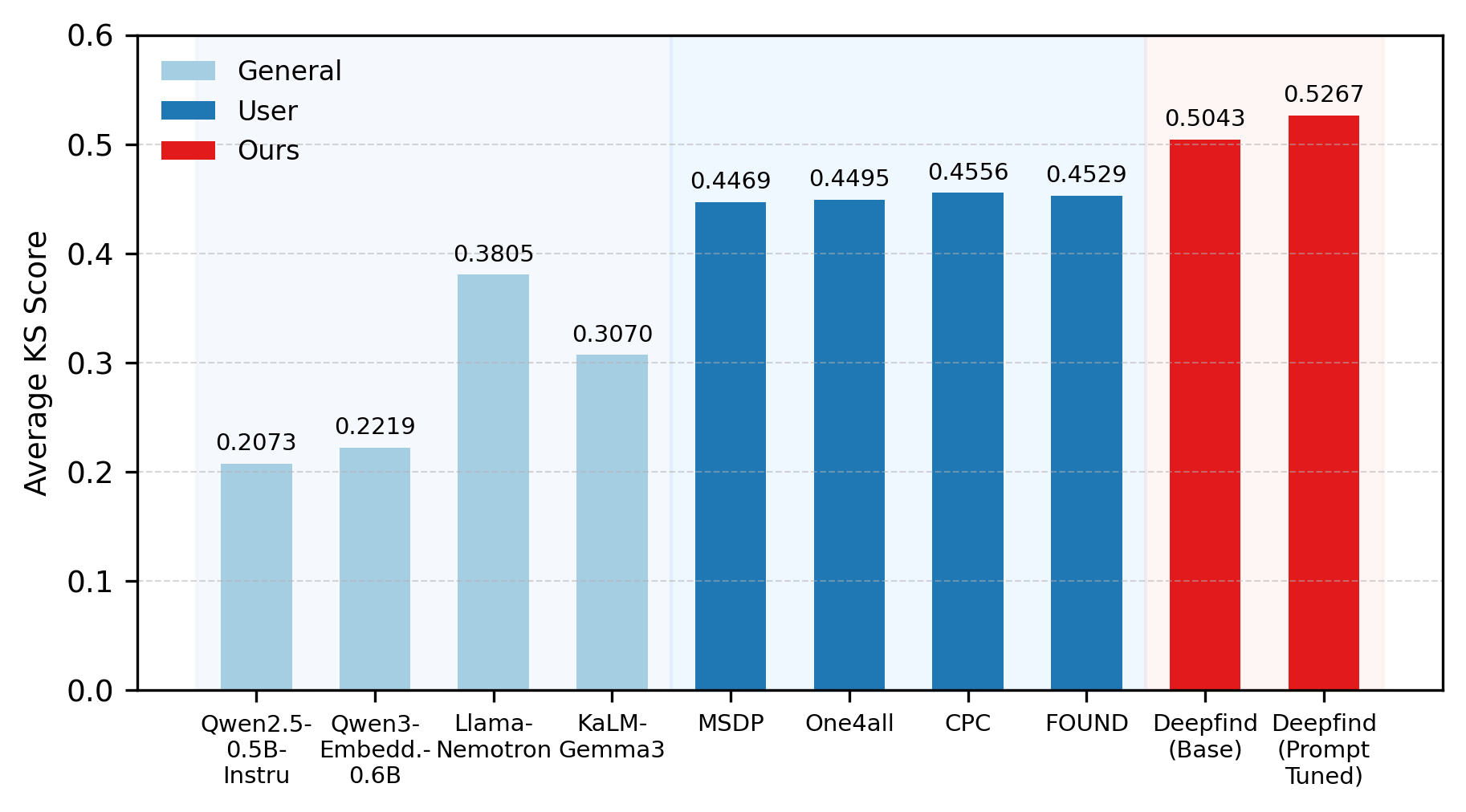} 
\caption{Average KS performance of Q-Anchor and baselines across 10 Alipay scenarios. Per-scenario results are provided in Appendix~\ref{subapp:ks}.}
\label{fig:ks}
\end{figure}

\textit{\textbf{\textsc{Q-Anchor} delivers the strongest and most stable AUC and KS across all scenarios.}}
As shown in Table~\ref{tab:results} and Fig.~\ref{fig:ks}, \textit{Q-Anchor (Prompt Tuned)} achieves the best performance across all 10 benchmarks, with an average AUC of 0.8225 and KS of 0.5267. It surpasses the strongest general-purpose baseline, Llama-Embed-Nemo- tron-8B (AUC: 0.7488, KS: 0.3805), by +0.0737 (+9.84\%) in AUC and +0.1462 (+38.4\%) in KS, consistently across User Engagement, Risk, and Marketing domains (see Appendix~\ref{subapp:ks}). These results indicate that the key limitation in industrial user modeling lies not in semantic capacity, but in representation alignment: generic text embeddings struggle to reconcile sparse, symbolic, multi-source behavioral logs. By contrast, \textsc{Q-Anchor}, pretrained on \textit{UserU} with hierarchical behavior encoding, maps heterogeneous events into query-relevant signals without relying on massive parameterization.

\textit{\textbf{Robust cross-domain representation validates Query-as- Anchor as a “one-model-for-many” paradigm.}}
The same encoder generalizes well across three heterogeneous domains-- Engagement, Risk, and Marketing—without task-specific architectures. In Risk, \textit{Q-Anchor (Prompt Tuned)} achieves 0.9439 on \textit{Money}, outperforming strong user-embedding baselines such as \textit{FOUND} (0.9235) and \textit{MSDP} (0.8746), suggesting that the anchor query suppresses high-entropy transactional noise and amplifies decision-relevant patterns. In Marketing, the gains are even more pronounced: for \textit{Brand}, AUC jumps from 0.7979 (Base) to 0.8535 (Prompt Tuned), indicating that the model not only transfers across domains but also adapts to domain-specific decision boundaries where subtle preference signals matter. This supports a scalable alternative to maintaining many scenario-specific models: a universal representation plus lightweight scenario conditioning.

\begin{figure}[t]
    \centering
    \begin{subfigure}{0.4955\columnwidth} 
        \centering
        \includegraphics[width=\linewidth]{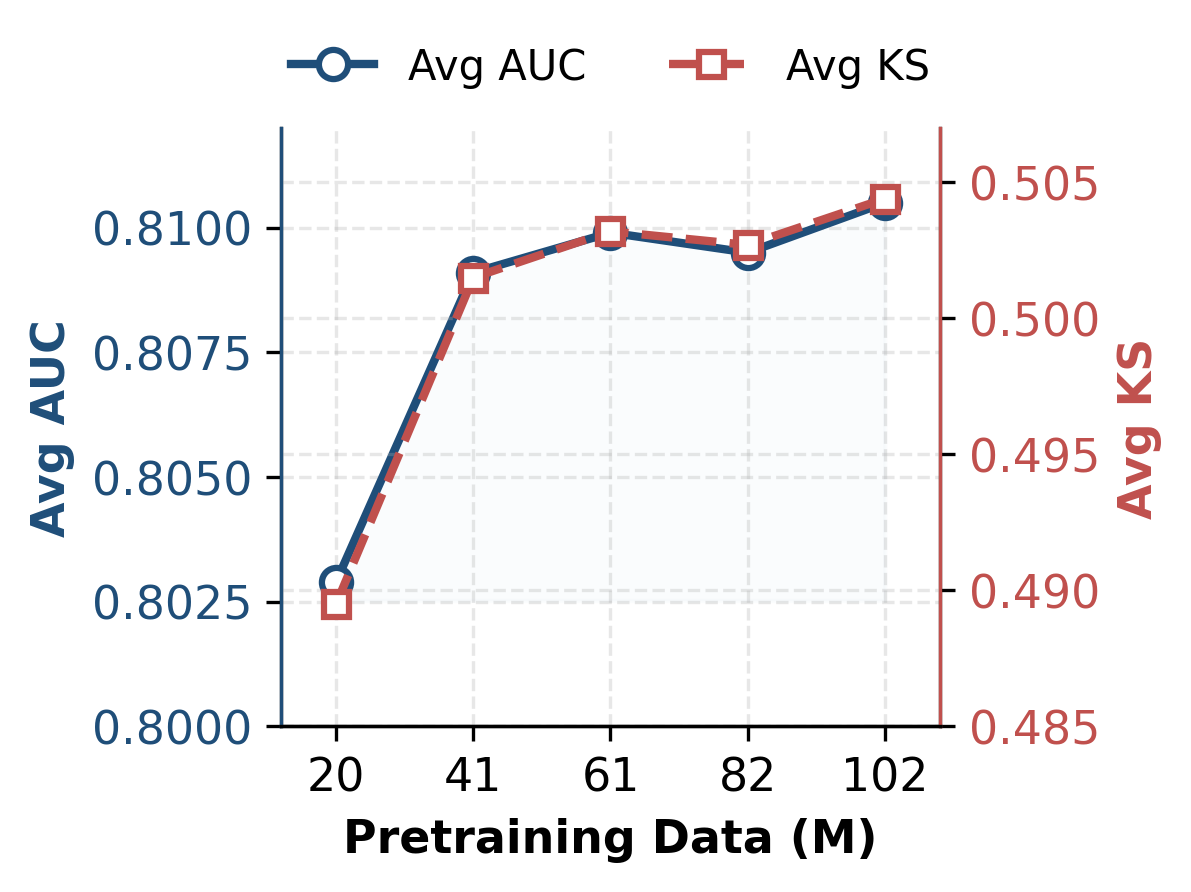}
        \caption{Pretraining Data Scale}
        \label{fig:data_scale}
    \end{subfigure}
    \hfill
    \begin{subfigure}{0.4955\columnwidth}
        \centering
        \includegraphics[width=\linewidth]{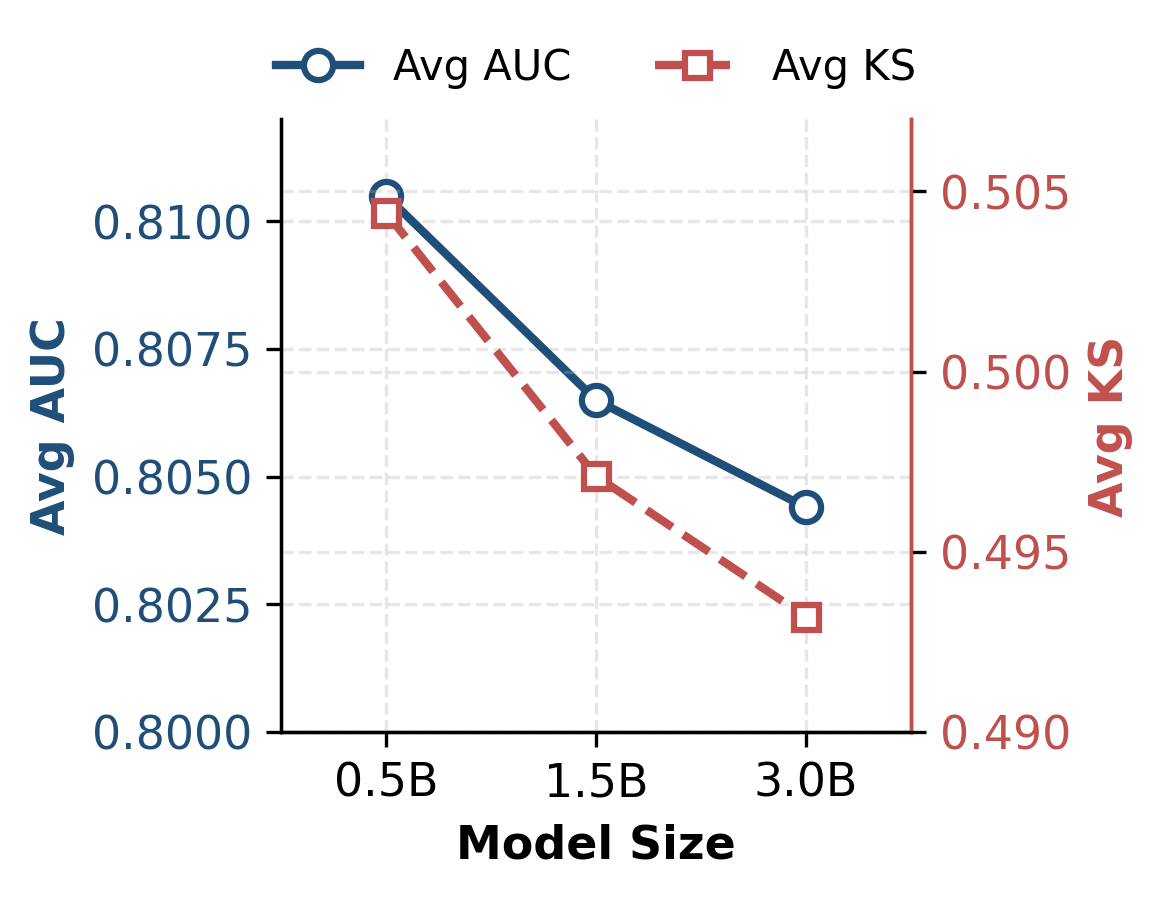}
        \caption{Impact of Model Size}
        \label{fig:model_scale}
    \end{subfigure}
    \caption{Scalability analysis of Q-Anchor Embedding (Base). (a) Performance increases with more pretraining data. (b) Performance vs. model parameters (0.5B to 3.0B).}
    \label{fig:base_scalability}
\end{figure}

\textit{\textbf{\textsc{Q-Anchor} (Base) scales more with data than with parameters.}}
Figure~\ref{fig:base_scalability} characterizes \textit{Q-Anchor (Base)} under \textbf{data scaling} (20.48M$\rightarrow$102.4M pretraining pairs) and \textbf{model scaling} (Qwen2.5-0.5B$\rightarrow$3B) with a fixed training/data budget. Increasing data yields consistent gains (Avg. AUC 0.8029$\rightarrow$\textbf{0.8105}, Avg. KS 0.4895$\rightarrow$\textbf{0.5044}). In contrast, model scaling is non-monotonic: Table~\ref{tab:model_scaling_base} shows \textbf{0.5B} performs best (Avg. AUC/KS=\textbf{0.8105}/\textbf{0.5044}), while 1.5B/3B bring no gains and sometimes regress (e.g., \textit{Brand}, \textit{Promo}), consistent with prior embedding-scaling observations~\cite{jiang2024scaling}. To better understand this, we analyze larger models' training gradient in Appendix~\ref{subapp:per_scenario_base} (Fig.~\ref{fig:gradient_scale}), showing that embedding quality depends more on pretraining data than model scale. Accordingly, we adopt 50k-step pretraining with the 0.5B backbone for the optimal accuracy–efficiency trade-off; per-scenario results are provided in Appendix~\ref{subapp:per_scenario_base}.

\begin{figure}[t]
    \centering
    \begin{subfigure}{0.4955\columnwidth}
        \centering
        \includegraphics[width=\linewidth]{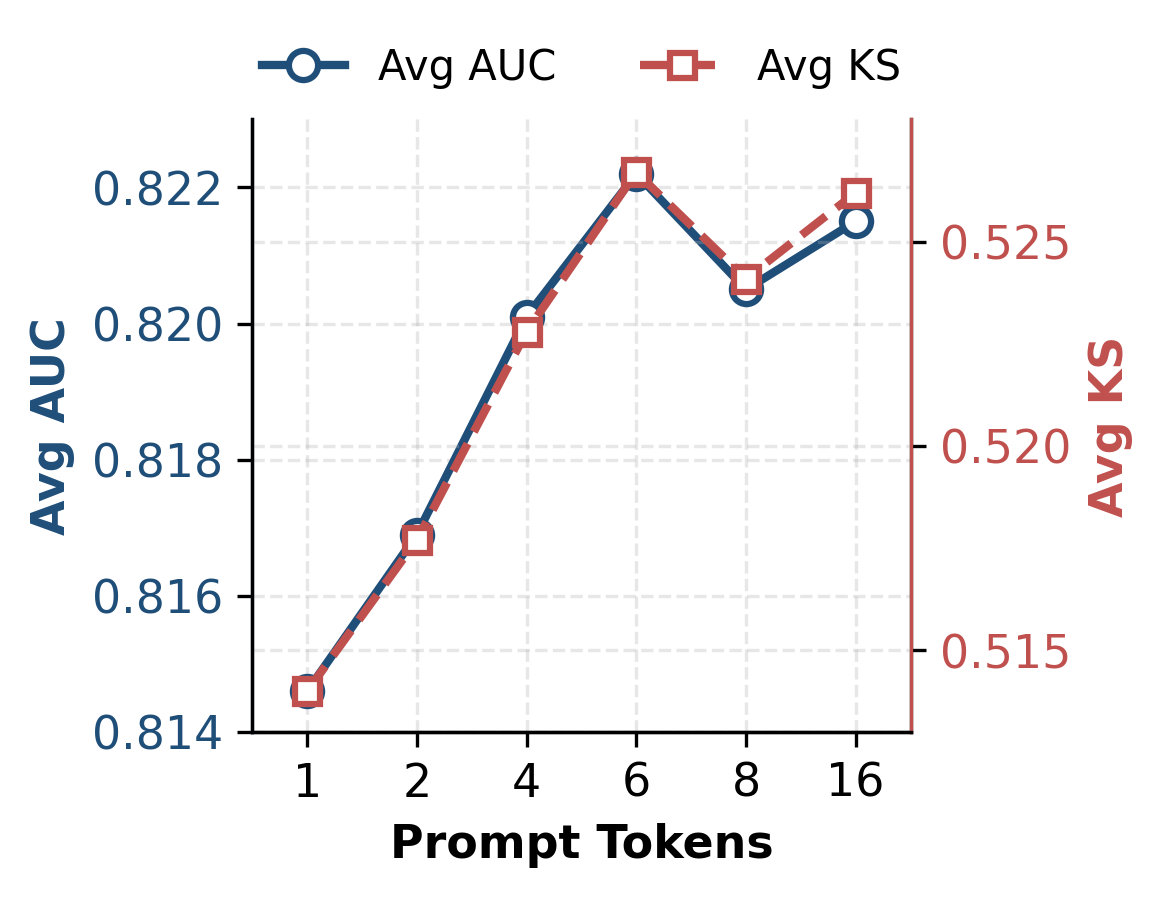}
        \caption{Prompt Tokens Scale}
        \label{fig:pt_token_scale}
    \end{subfigure}
    \hfill
    \begin{subfigure}{0.4955\columnwidth}
        \centering
        \includegraphics[width=\linewidth]{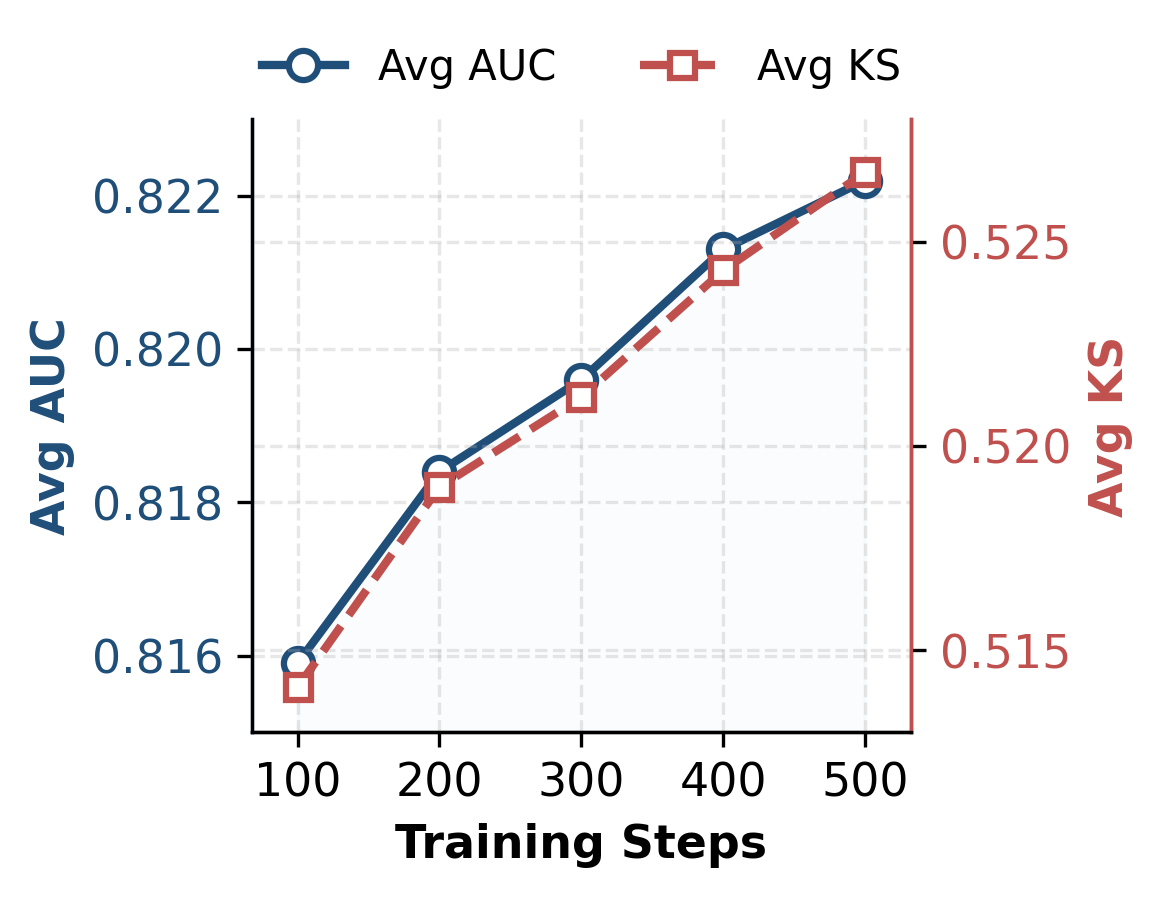}
        \caption{Training Steps Scale}
        \label{fig:pt_step_scale}
    \end{subfigure}
    \caption{Scalability analysis of Q-Anchor Embedding (Prompt Tuned). (a) Performance vs.\ the number of learnable prompt tokens (1-16, Ours: 6). (b) Performance vs.\ the training-step budget (100--500, Ours: 500).}
    \label{fig:pt_scalability}
\end{figure}

\textit{\textbf{\textsc{Q-Anchor} (Prompt-tuned) scales efficiently with prompt tokens and tuning steps, saturating early.}}
Figure~\ref{fig:pt_scalability} evaluates \textit{Q-Anchor Embedding (Prompt Tuned)} under \textbf{token scaling} (1$\rightarrow$16 tokens) and \textbf{step scaling} (100$\rightarrow$500 steps). Performance rises rapidly up to \textbf{6 tokens} (Avg. AUC 0.8146$\rightarrow$\textbf{0.8225}, Avg. KS 0.5140$\rightarrow$\textbf{0.5267}), then saturates with minor fluctuations at 8/16 tokens. Increasing tuning steps yields steady gains (Avg. AUC 0.8159$\rightarrow$\textbf{0.8225}, Avg. KS 0.5141$\rightarrow$\textbf{0.5267}). Overall, prompt tuning is \textbf{efficiency-friendly}, achieving most gains with a small prompt budget and modest optimization, motivating \textbf{6 tokens} and \textbf{500 steps} as our default. Per-scenario results are in Appendix~\ref{subapp:per_scenario_pt}.

\begin{figure}[t]
    \centering
    \includegraphics[width=0.495\linewidth]{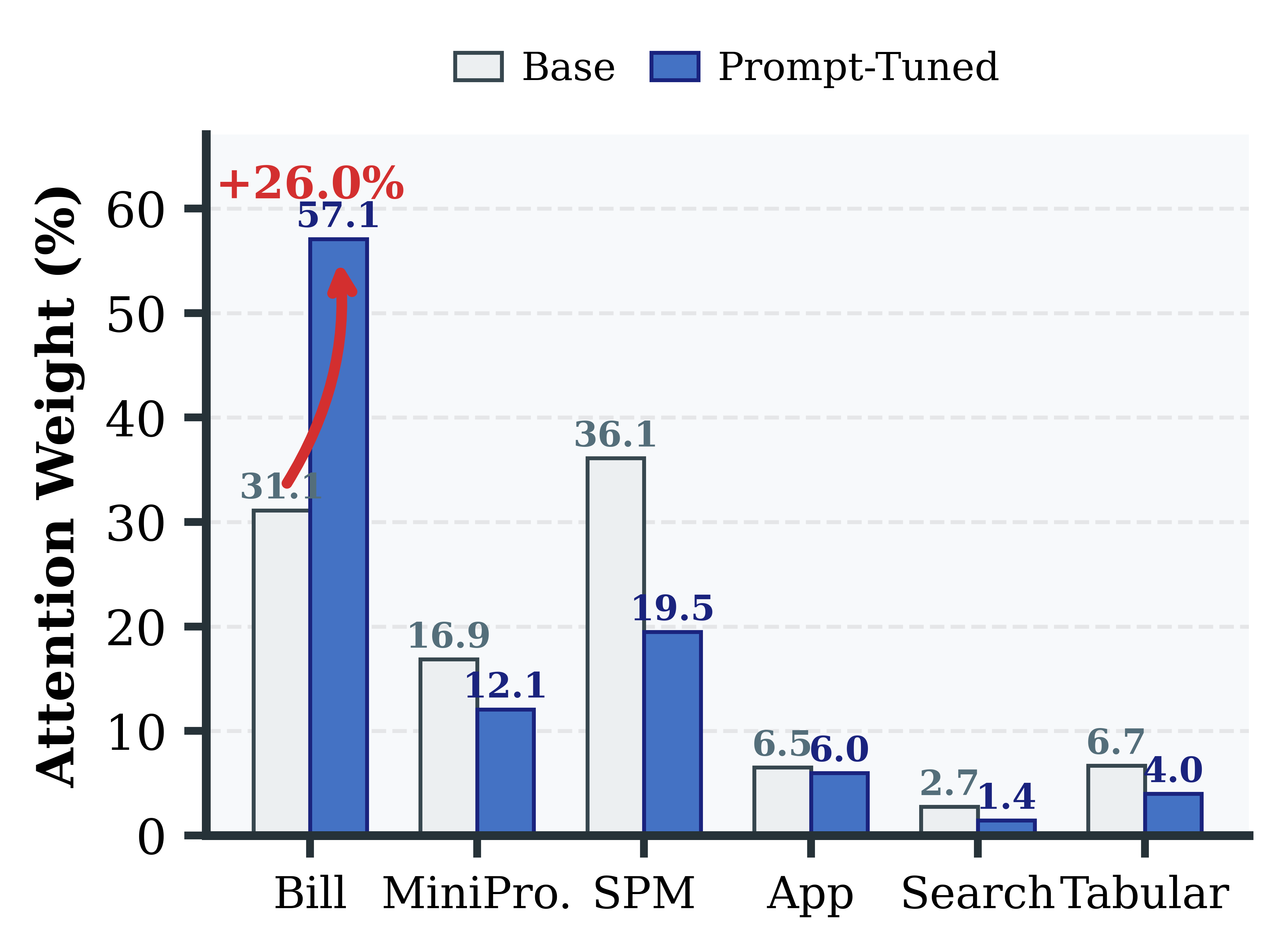}
    \hfill
    \includegraphics[width=0.495\linewidth]{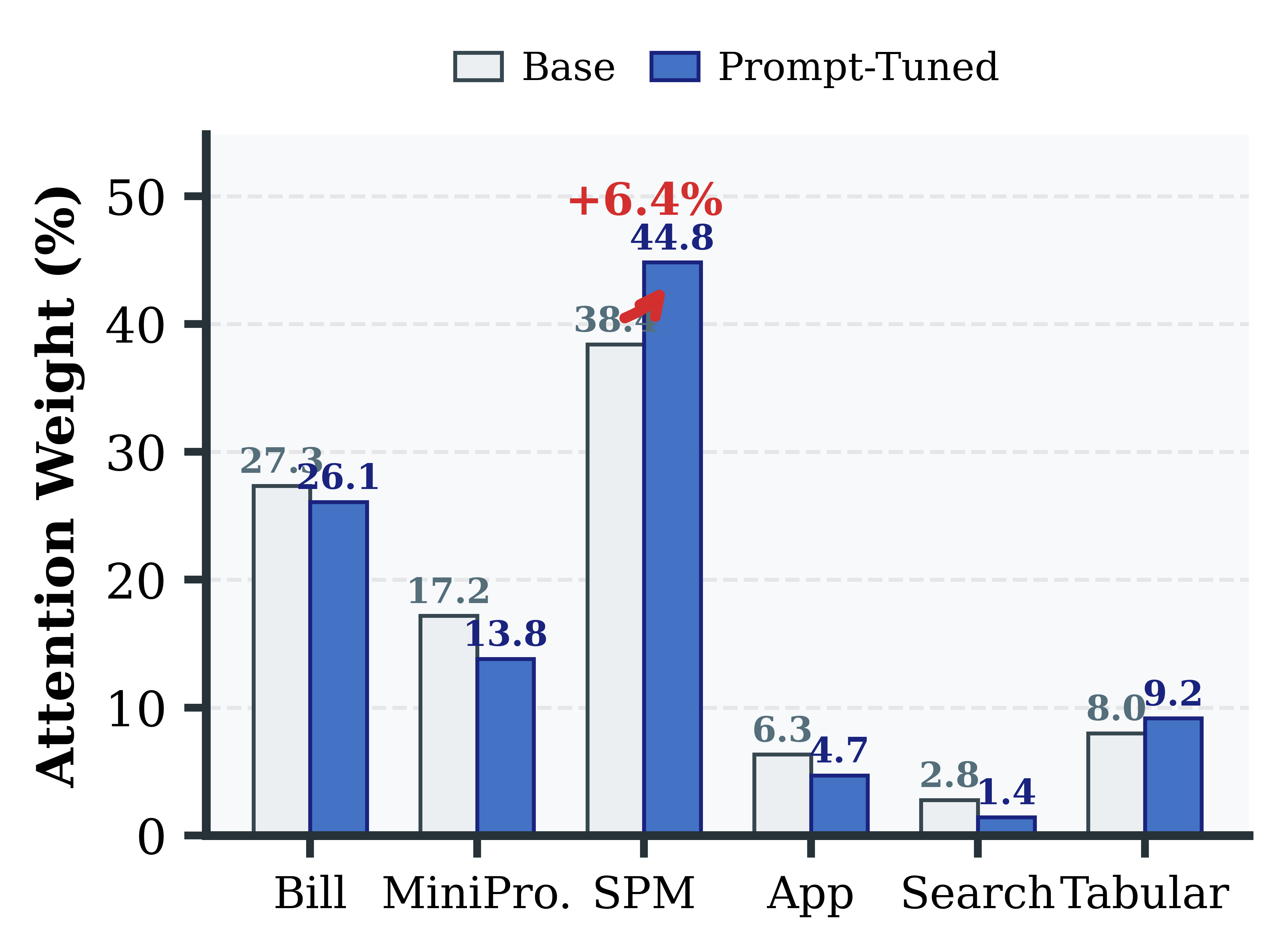}
    \caption{
Modal attention shift after prompt tuning. \textbf{Left}: Takeout Interest—Bill attention increases by +26.0\%, capturing dining purchasing power. \textbf{Right}: Ant Forest—SPM attention rises by +6.4\%, consistent with navigation-heavy usage. Red arrows mark positive attention gains, highlighting Q-Anchor’s dynamic feature re-anchoring.
    }
    \label{fig:attention_shift_total}
\end{figure}

\begin{figure}[htbp]
\captionsetup[subfigure]{skip=1pt}
    \centering
    \begin{subfigure}[b]{0.9\linewidth}
        \centering
        \includegraphics[width=\linewidth]{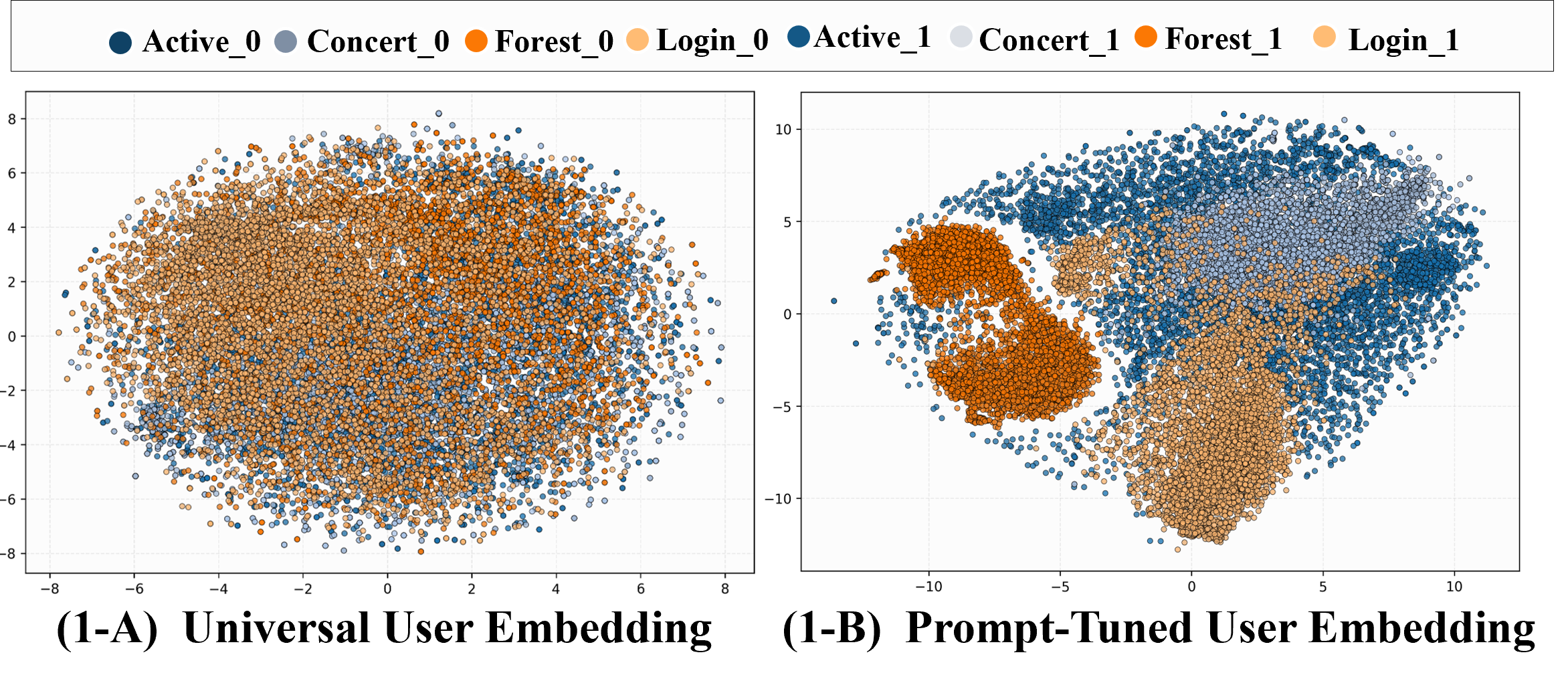}
        \caption{User Engagement}
        \label{fig:1}
    \end{subfigure}
    % \hfill    
    \begin{subfigure}[b]{0.9\linewidth}
        \centering
        \includegraphics[width=\linewidth]{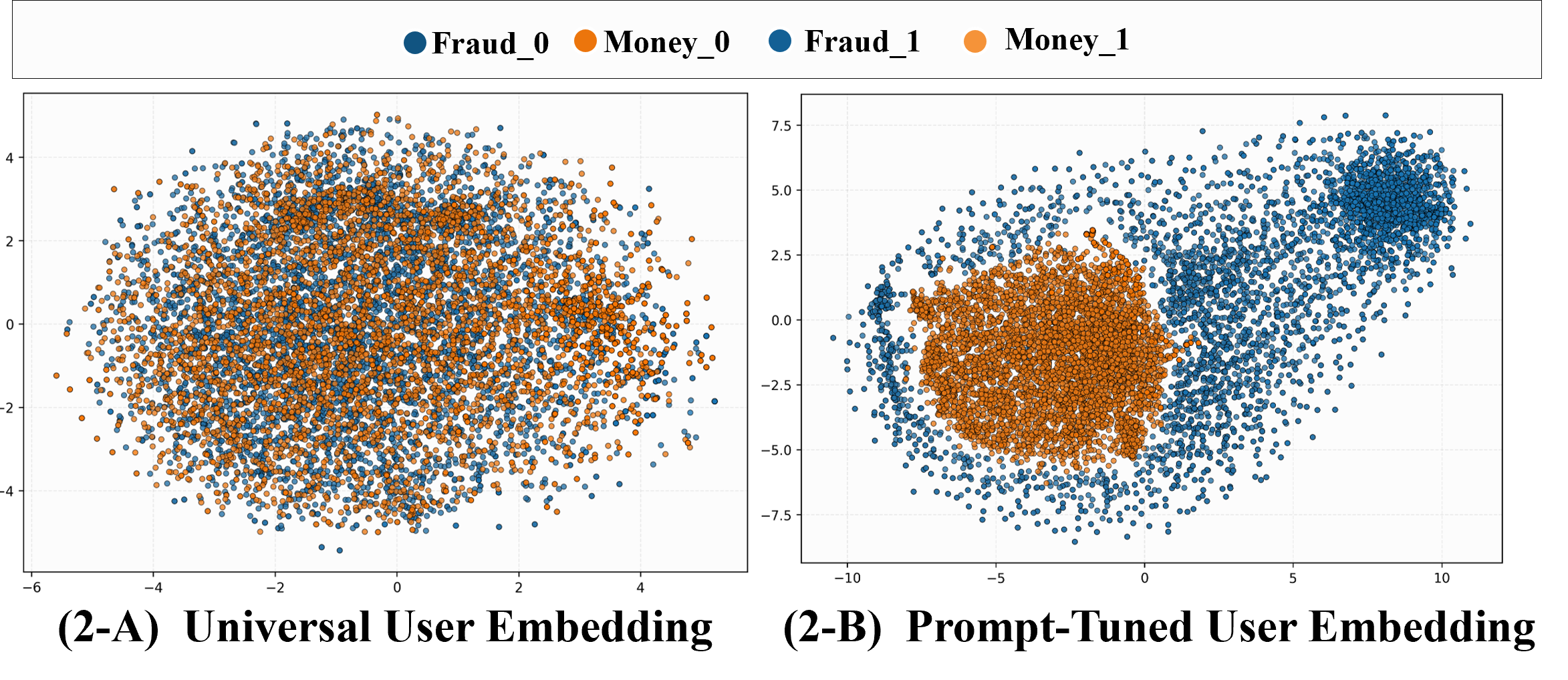}
        \caption{Risk}
        \label{fig:2}
    \end{subfigure}
    
    % \hfill    
    
    \begin{subfigure}[b]{0.9\linewidth}
        \centering
        \includegraphics[width=\linewidth]{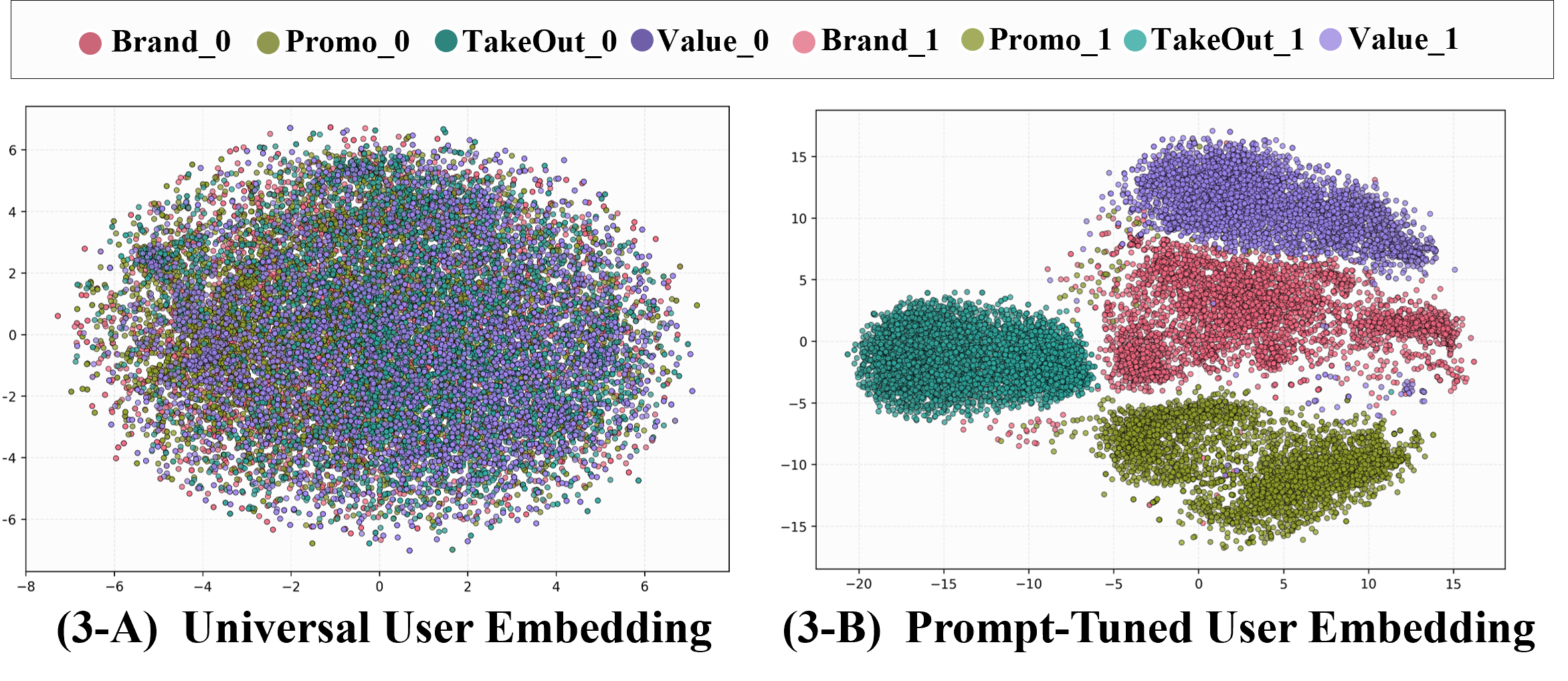}
        \caption{Marketing Sensitivity}
        \label{fig:3}
    \end{subfigure}
    
    \caption{t-SNE Visualization of universal and prompt-tuned representation of 10 scenarios. PCA visualizations are provided in Appendix~\ref{subapp:PCA}.}
    \label{fig:tsne}
\end{figure}

\textit{\textbf{Scenario-adaptive tuning as a performance multiplier with interpretable re-anchoring.}}
Post-training with cluster-based soft prompts yields a consistent lift in \emph{all} scenarios (avg. 0.8104 to 0.8225), confirming that the base model captures general behavioral semantics while prompts specialize the embedding space to scenario boundaries. Importantly, the improvement is not a black box: Figure~\ref{fig:attention_shift_total} shows that prompt tuning changes \emph{where} the model looks. For \textit{Takeout Interest}, attention to the \textit{Bill} modality increases by +26.0\%, aligning with purchasing power as the dominant signal; for \textit{Ant Forest}, the \textit{SPM} modality rises by +6.4\%, matching navigation- \\intensive usage. The t-SNE visualizations (Figure~\ref{fig:tsne}) further corroborate this behavior: prompt-tuned representations form clearer \\ scenario-consistent groupings than the universal space, reflecting better separation of positives vs. negatives. To avoid visualization bias, we corroborate t-SNE findings with PCA (Appendix~\ref{subapp:PCA}), which consistently shows sharper cluster separation under prompt tuning. Together, the quantitative gains and qualitative shifts indicate that Query-as-Anchor functions as a semantic lens—guiding the representation to focus more on semantically relevant multi-modal evidence through minimal parameter updates—enabling fast, deployable specialization in changing industrial environments.

\subsection{Ablation Study}
We ablate \textit{Q-Anchor (Base)} by removing one component at a time while keeping the backbone, data, and training budget fixed. We evaluate (i) structural tokens (User token, modality tokens, or both), (ii) pretraining ingredients (margin-mask filtering, NTP, and contrastive alignment), and (iii) post-training (prompt tuning, and prompt tuning without pretraining). AUC results are in Table~\ref{tab:ablations}, with KS results in Appendix~\ref{subapp:ks}.

\begin{table*}[htbp]
\centering
\footnotesize
\caption{Ablation study (AUC) of Q-Anchor Embedding across modality, training method and prompt tuning. KS results are detailed in Appendix~\ref{subapp:ks}.}
\label{tab:ablations}
\begin{tabular*}{\textwidth}{@{\extracolsep{\fill}} l *{11}{c} @{}}
\toprule
& \multicolumn{4}{c}{\textbf{User Engagement}} 
& \multicolumn{2}{c}{\textbf{Risk}} 
& \multicolumn{4}{c}{\textbf{Marketing Sensitivity}} 
& \textbf{Avg. AUC} \\
\cmidrule(lr){2-5} \cmidrule(lr){6-7} \cmidrule(lr){8-11}

\textbf{Method} 
& \textbf{Active} 
& \textbf{Concert} 
& \textbf{Login} 
& \textbf{Forest} 
& \textbf{Fraud} 
& \textbf{Money} 
& \textbf{Takeout} 
& \textbf{Brand} 
& \textbf{Promo} 
& \textbf{Value}
& \\
\midrule
\textsc{Q-Anchor} (Base)    
& \underline{0.6568} & \underline{0.5739} & 0.8420 & \underline{0.9700} & \underline{0.9218} & \textbf{0.9382} & \textbf{0.8799} & \textbf{0.7979} & 0.6189 & \textbf{0.9049} & \textbf{0.8104} \\
\midrule
\multicolumn{12}{@{}l@{}}{\textit{\textbf{Ablation on Modality Token}}} \\
w / o User Tok.
& 0.6525	& \textbf{0.5757}	& \underline{0.8424}	& \textbf{0.9702}	& \textbf{0.9219}	& 0.9358	& \underline{0.8785}	& 0.7884	& \underline{0.6195}	& 0.9006	& 0.8086
\\

w / o Modal Tok.
& 0.6546	& 0.5701	& 0.8423	& 0.9700	& 0.9214	& 0.9372	& 0.8778	& 0.7903	& \textbf{0.6210}	& \underline{0.9036}	& \underline{0.8088}
\\
w / o Modal \& User Tok.
& 0.6544	& 0.5703	& 0.8403	& 0.9694	& 0.9211	& \underline{0.9374}	& 0.8783	& \underline{0.7819}	& 0.6141	&  0.8981 & 0.8065 \\

\midrule
\multicolumn{12}{@{}l@{}}{\textit{\textbf{Ablation on Training Method}}} \\
w / o Filter 
& 0.6461	
& 0.5693	
& 0.8412	
& 0.9685
& 0.9197	
& 0.9357	
& 0.8728	
& 0.7867	
& 0.6218	
& 0.8854	
& 0.8047 \\

w / o NTP
& \textbf{0.6577}
& 0.5692
& 0.8392
& 0.9701
& 0.9208
& 0.9369
& 0.8798
& 0.7858
& 0.6079
& 0.8936
& 0.8061
\\

w / o Contrastive
& 0.6408	
& 0.5456	
& \textbf{0.8454}	
& 0.9584	
& 0.9071	
& 0.9091	
& 0.7967	
& 0.6512	
& 0.5980
& 0.8143
& 0.7667
\\
% \bottomrule
\midrule
% \midrule
\textsc{Q-Anchor} (Prompt Tuned)      
& \textbf{0.6678} & \textbf{0.5844} & \textbf{0.8443} & \textbf{0.9716} & \textbf{0.9242} & \textbf{0.9439} & \textbf{0.8811} & \textbf{0.8535} & \textbf{0.6350} & \textbf{0.9194} & \textbf{0.8225} \\
\midrule
\multicolumn{12}{@{}l@{}}{\textit{\textbf{Ablation on Prompt Tuning}}} \\
w / o pretrain
& 0.6557
& 0.5432
& 0.7546
& 0.9338
& 0.9189
& 0.8153
& 0.8793
& 0.7937
& 0.5833
& 0.9044
& 0.7782 \\

\bottomrule
\end{tabular*}
\end{table*}
\textit{\textbf{Ablation on Modality Token: } \textbf{Explicit structure helps Deep-Find attribute evidence to the correct source.}}
Removing the \textit{User} or modality tokens slightly lowers Avg. AUC (0.8104 to 0.8086/ 0.8088), and removing both produces the largest drop in this block (0.8065). The effect concentrates on modality-sensitive marketing scenarios, especially \textit{Brand} (0.7979 to 0.7819), where performance depends on \emph{which} modality carries the signal rather than overall activity. These results support our design: user/modal markers inject a minimal inductive bias about log structure and improve cross-modal aggregation; consistent degradations are also observed in KS (Appendix~\ref{subapp:ks}), indicating reduced ranking separability.

\textit{\textbf{Ablation on Training Method: } \textbf{Contrastive alignment is the primary signal; auxiliary objectives act as regularizers.}}
Removing contrastive learning causes the largest regression, reducing Avg. AUC from 0.8104 to 0.7667 and sharply degrading fine-grained scenarios (e.g., \textit{Brand} 0.7979$\rightarrow$0.6512; \textit{Money} 0.9382$\rightarrow$0.9091). This indicates that token-level modeling alone cannot impose query-conditioned separability, whereas contrastive supervision explicitly structures the embedding space. In comparison, margin-mask filtering and next-token prediction are \emph{supportive}: disabling filtering (``w/o Filter'') lowers Avg. AUC to 0.8047, consistent with increased false negatives in noisy logs, and disabling NTP (``w/o NTP'') yields 0.8061, suggesting weaker local event modeling. KS shows an even steeper drop without contrastive alignment (Appendix~\ref{subapp:ks}), and the same trend is reflected qualitatively by less coherent scenario clusters in Fig.~\ref{fig:tsne}. Overall, contrastive alignment defines the geometry, while filtering and NTP mainly stabilize training and improve robustness.

\textit{\textbf{Ablation on Pretraining:} 
\textbf{Pre-trained weights provide the essential behavioral prior for intent discovery.} }
Excluding the pretraining phase (\textit{w/o pretrain}) triggers a systemic performance collapse: Average AUC falls from \textbf{0.8225} to \textbf{0.7781}, while Average KS undergoes a sharper relative regression of 11.2\% (\textbf{0.5267}$\rightarrow$\textbf{0.4679}). The degradation is most pronounced in complex scenarios (e.g., Money AUC: \textbf{0.9439}$\rightarrow$\textbf{0.8153}), where prompt tuning alone fails to reconstruct the high-dimensional latent structures necessary for fine-grained separation. This confirms that pretraining is not a mere optimization aid but a foundational requirement for distilling robust behavioral priors, without which the model cannot effectively generalize across noisy, long-tail user trajectories.

\vspace{-1em}
\subsection{Industrial Online A/B Testing}
We evaluate \textsc{Q-Anchor} embeddings in two large-scale Alipay A/B tests over \textbf{two weeks}, with users randomly assigned to treatment (policy with embeddings) or control (fixed-time/rule-based). See Appendix~\ref{app:offline} for pre-deployment offline results; online A/B outcomes are reported below:

\textit{\textbf{Scenario I: Interactive Voice Response (IVR) Cash-Reserve Outreach.}}  
\textsc{Q-Anchor} embeddings were deployed in the live cash-reserve outreach system to optimize send-time decisions according to user availability and responsiveness. In production, this represen- tation-aware timing increased the \textbf{drawdown rate by 12.5\%} and the \textbf{average outstanding balance per user by 5.3\%}. Early-funnel engagement also improved: the \textbf{cash-reserve product visit rate rose by 4.2\%}, and \textbf{drawdown-page visits increased by 17.7\%}, reflecting stronger activation and smoother progression through the credit funnel. These results demonstrate that embedding-informed timing decisions enhance both immediate credit utilization and upstream engagement.

\textit{\textbf{Scenario II: Credit Delinquency Risk Identification.}}  
\textsc{Q-Anchor} embeddings were integrated into the credit risk scoring pipeline, improving the business-critical \textbf{KS score by 1.96\%}. Results show that embedding-informed scoring enhances predictive performance and risk-aware credit allocation.

\textit{\textbf{Deployment at Alipay Scale.}}
We run daily \textsc{Q-Anchor embedding} refresh for hundreds of millions of users on a \textbf{100$\times$L20} cluster. For multi-scenario serving, \textbf{Query-as-Anchor} with shared \textbf{prefix KV-cache} encodes each user prefix once and reuses it across queries, so adding a scenario only requires query-suffix computation (\textbf{one extra L20} to maintain the same SLA), rather than replicating the entire 100-GPU pipeline for every scenario.

\section{Conclusion}
We present \textsc{Q-Anchor Embedding}, a unified framework for industrial user representation that closes the gap between sparse, noisy, heterogeneous behavior logs and LLM-level semantic representatin. \textsc{Q-Anchor} contributes: (1) \textbf{UserU}, an industrial-scale pretraining corpus that couples rule-based future behavior supervision with reflection-verified LLM-synthesized user QA to inject temporal dynamics and semantic understanding; (2) Query-as-\\Anchor, a hierarchical coarse-to-fine user encoder and query- conditioned alignment paradigm that distills multi-source signals into task-adaptive embeddings; and (3) \textbf{semantic re-anchoring} via lightweight soft prompts that re-anchor modalities to match downstream decision boundaries. Across 10 real-world benchmarks and 2 online A/B tests, \textsc{Q-Anchor} achieves SOTA AUC and KS over LLM embeddings and strong baselines, while enabling interpretable, scenario-adaptive, low-cost, and transferable industrial user representations.
\section*{Ethical Considerations}
All experiments comply with Alipay's data-governance, privacy, and security policies, with encrypted data/embeddings, strict access control, and audit logging to prevent unauthorized access or linkage. We present \textsc{Q-Anchor Embedding} to advance responsible user representation learning and industrial systems.
\clearpage
\bibliographystyle{ACM-Reference-Format}
\bibliography{custom}

% %%
% %% If your work has an appendix, this is the place to put it.
\appendix
\section{Data Example of \textsc{UserU} Dataset}
\label{app:dataexample}
In this section, we present \textbf{illustrative toy examples} curated from the two core subsets of the \textsc{UserU} dataset: the \textbf{Behavior-based Interaction Dataset ($\mathcal{D}_{\text{future}}$)} and the \textbf{Synthetic Query-Answer Dataset ($\mathcal{D}_{\text{uqa}}$)}. These samples are designed to provide a qualitative intuition of the data modality and the alignment tasks.

\textbf{Privacy and Demonstration Note:} It is important to emphasize that the cases provided below are \textbf{synthetic toy examples constructed solely for demonstration purposes}. They do not represent the actual raw records of any specific individual.
\begin{figure}[htbp]
    \centering

    % -------- Subfigure (a) --------
    \begin{subfigure}{\linewidth}
        \centering
        \includegraphics[width=\linewidth]{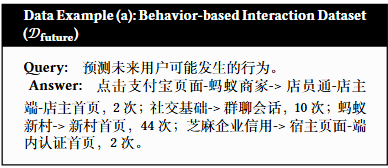}
        \caption{Alignment between raw multi-modal behavioral signals and future user trajectories.}
        \label{fig:data_future}
    \end{subfigure}

    \vspace{2em}

    % -------- Subfigure (b) --------
    \begin{subfigure}{\linewidth}
        \centering
        \includegraphics[width=\linewidth]{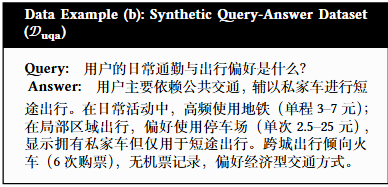}
        \caption{High-level semantic reasoning derived from implicit behavioral patterns.}
        \label{fig:data_uqa}
    \end{subfigure}

    \vspace{1em}
    \caption{Illustrative toy instances from the \textsc{UserU} dataset. Figure (a) illustrates the behavior-to-behavior prediction task, while Figure (b) showcases the behavior-to-semantic reasoning capability enabled by query-aware alignment.}
    \label{fig:combined_data_examples}
\end{figure}

\section{Details of Downstream Benchmarks}
\label{app:dataset_details}
As shown in Table~\ref{tab:task} \& Table~\ref{tab:task_definitions}, we evaluate on $\mathcal{D}_{test}$ containing 10 real-world binary classification tasks, grouped into three domains.

\paragraph{Common protocol.}
Each benchmark is a user-level binary classification task.
For task $\tau$, we define a prediction window and assign the label
$y_u^\tau \in \{0,1\}$ by whether user $u$ triggers at least one target event
associated with $\tau$ during that window.
We report task definitions at the scenario level for privacy and compliance;
all methods are evaluated under identical data construction and labeling rules.

\begin{table*}[!h]
\centering
\small
\renewcommand{\arraystretch}{1}
\caption{Downstream test benchmarks and scenario-level binary label definitions. Labels are assigned by whether a user triggers the target event within the prediction window. Each task contains $\approx$50w test data.}
\label{tab:task_definitions}

\resizebox{\textwidth}{!}{%
\begin{tabular}{l p{0.2\textwidth} p{0.2\textwidth} p{0.28\textwidth} p{0.26\textwidth}}
\hline
\textbf{Domain} & \textbf{Scenario (Task)} & \textbf{Target} & \textbf{Positive ($y{=}1$)} & \textbf{Negative ($y{=}0$)} \\
\hline

\multirow{4}{*}{User Engagement}
& Interest Community Active User Identification (Active)
& Whether a user will be active in the Alipay interest community.
& Any qualified interest-community activity occurs (e.g., community visit/app open, content browse, like, comment, post).
& No qualified interest-community activity occurs. \\

& Concert Click Prediction (Concert)
& Whether a user will click concert-related content.
& Any click on items categorized as concert-related occurs.
& No click on concert-related items occurs. \\

& User Log-in Prediction (Login)
& Whether a user will log in.
& At least one login event occurs.
& No login event occurs. \\

& Ant Forest Engagement (Forest)
& Whether a user will engage with Ant Forest.
& Any Ant Forest interaction occurs (e.g., enter/collect/participate).
& No Ant Forest interaction occurs. \\
\hline

\multirow{2}{*}{Risk}
& Fraud Detection (Fraud)
& Whether a user will be associated with a fraud-related risk event.
& User is confirmed/flagged by the risk-control pipeline as fraud-related.
& User is not flagged as fraud-related. \\

& Money Laundering Detection (Money)
& Whether a user will be associated with a money-laundering-related risk event.
& User is confirmed/flagged by compliance/risk-control as money-laundering-related.
& User is not flagged as money-laundering-related. \\
\hline

\multirow{4}{*}{Marketing Sensitivity}
& Takeout Interest (Takeout)
& Whether a user will show interest in takeout-related services.
& Any click/visit/conversion on takeout-category content occurs.
& No click/visit/conversion on takeout-category content occurs. \\

& Brand Sensitivity (Brand)
& Whether a user will respond to brand-related marketing stimuli.
& Any interaction with brand campaigns occurs (e.g., click/claim/participate).
& No interaction with brand campaigns occurs. \\

& Big Sale Sensitivity (Promo)
& Whether a user will respond to big-sale promotions.
& Any interaction with big-sale promotional content occurs.
& No interaction with big-sale promotional content occurs. \\

& Cost-Performance Sensitivity (Value)
& Whether a user will respond to cost-performance offers.
& Any interaction with value-for-money themed content or items tagged as cost-performance occurs.
& No such interaction occurs. \\
\hline

\end{tabular}%
}
\end{table*}

\section{Supplementary Experiments and Analysis}
\label{app:add}
\subsection{Experimental Results on Alipay Benchmarks: KS Performance and Ablation Study}
\label{subapp:ks}
\paragraph{\textbf{KS Performance.}} As illustrated in Table~\ref{tab:ksresults}, our Q-Anchor embedding achieves state-of-the-art KS performance across all 10 industrial scenarios, demonstrating superior discriminative power in ranking positive vs. negative user behaviors. As shown in Table~\ref{tab:results}, Q-Anchor (Prompt Tuned) attains the highest average KS (0.5267), outperforming not only general-purpose embedding models (e.g., Llama-Embed-Nemotron-8B: 0.3805) but also specialized user representation methods such as FOUND (0.4529) and CPC (0.4556).
\begin{table*}[htbp]
\centering
\small
\caption{KS performance on 10 key scenarios where our Q-Anchor (base version: pretrained) and (prompt-tuning version: post-trained) shows consistent improvement.}
\label{tab:ksresults}
\begin{tabular*}{\textwidth}{@{\extracolsep{\fill}} l *{11}{c} @{}}
\toprule
& \multicolumn{4}{c}{\textbf{User Engagement}} 
& \multicolumn{2}{c}{\textbf{Risk}} 
& \multicolumn{4}{c}{\textbf{Marketing Sensitivity}} 
& \textbf{Avg. KS} \\
\cmidrule(lr){2-5} \cmidrule(lr){6-7} \cmidrule(lr){8-11}

\textbf{Method} 
& \textbf{Active} 
& \textbf{Concert} 
& \textbf{Login} 
& \textbf{Forest} 
& \textbf{Fraud} 
& \textbf{Money} 
& \textbf{TakeOut} 
& \textbf{Brand} 
& \textbf{Promo} 
& \textbf{Value}
& \\
\midrule

\multicolumn{12}{@{}l@{}}{\textit{\textbf{General Embedding Models}}} \\
Qwen2.5-0.5B-Instruct
& 0.0421
& 0.0280
& 0.3326
& 0.3174
& 0.3034
& 0.2958
& 0.2066
& 0.1324
& 0.0880
& 0.3262
& 0.2073
 \\

Qwen3-Embedding-0.6B     
& 0.0524
& 0.0330
& 0.3459
& 0.3274
& 0.3102
& 0.3086
& 0.275
& 0.1581
& 0.0814
& 0.3270
& 0.2219

 \\
Llama-Embed-Nemotron-8B
& 0.1302
& 0.0857
& 0.4003
& 0.5201
& 0.5758
& 0.5868
& 0.4918
& 0.3685
& 0.1364
& 0.5091
& 0.3805
 \\

KaLM-Embed.-Gemma3-12B		
& 0.0880
& 0.0533
& 0.3830
& 0.4036
& 0.4944
& 0.4879
& 0.4207
& 0.2139
& 0.1312
& 0.3937
& 0.3070
\\

\midrule
\multicolumn{12}{@{}l@{}}{\textit{\textbf{User Embedding Models}}} \\
MSDP \cite{fu2023robust} 
& 0.2165
& 0.0608
& 0.7731
& 0.8104
& 0.7026
& 0.6069
& 0.4229
& 0.2107
& 0.1299
& 0.5345
& 0.4469

\\
One4all \cite{shin2021one4all} 
& \underline{0.2403}
& \underline{0.1075}
& \textbf{0.7749}
& 0.8191
&\textbf{ 0.7160}
& 0.6113
& 0.3886
& 0.2135
& 0.1261
& 0.4980
& 0.4495

 \\
CPC~\cite{oord2018representation} 
& \underline{0.2402}
& 0.0921
& \underline{0.7729}
& 0.8197
& 0.7077
& 0.6028
& 0.4249
& 0.2068
& 0.1824
& 0.5061
& 0.4556
 \\
FOUND \cite{dou2025transferable}
& 0.1669
& 0.0780
& 0.4660
& 0.7967
& 0.6739
& 0.7084
& 0.5762
& 0.3348
& \underline{0.1753}
& 0.5521
& 0.4529
  \\

\midrule
\multicolumn{12}{@{}l@{}}{\textit{\textbf{Q-Anchor Embedding (Ours)}}} \\
\textsc{Q-Anchor} (Base)    
& 0.2228 & 0.1023 & 0.5216	& \underline{0.8437} & 0.7054 & \underline{0.7450}&  \underline{0.6210} & \underline{0.4527} & 0.1727 & \underline{0.6567} & \underline{0.5044}  \\
\textsc{Q-Anchor} (Prompt Tuned)      
& \textbf{0.2412} & \textbf{0.1177} & 0.5225 & \textbf{0.8480} & \underline{0.7086} & \textbf{0.7573} & \textbf{0.6212} & \textbf{0.5564} & \textbf{0.1962} & \textbf{0.6978} & \textbf{0.5267} \\
\bottomrule
\end{tabular*}
\end{table*}

\paragraph{\textbf{KS Ablation Study.}}
Table~\ref{tab:ablations_ks} presents the KS ablations of \textit{Q-Anchor (Base)} by removing one component at a time while keeping the backbone, data, and training budget fixed. Overall, removing structural tokens (user/modality token) causes mild yet consistent degradations in Avg. KS, indicating that explicit log structure helps the encoder attribute evidence to the correct source and improves ranking stability. In contrast, the contrastive objective is the primary driver of discriminability: without contrastive learning, Avg. KS drops substantially from 0.5044 to 0.4215, with particularly large losses on fine-grained marketing scenarios (e.g., \textit{Brand}: 0.4527 $\rightarrow$ 0.2169), showing that token-level modeling alone is insufficient to enforce query-conditioned separation. Disabling margin-mask filtering or NTP yields smaller but systematic decreases, suggesting they mainly stabilize training under noisy, sparse behavioral logs. Notably, the "w/o pretrain" ablation reveals that skipping the initial pretraining phase significantly weakens the model's performance (Avg. KS drops by ~11.16\%), even with subsequent prompt tuning. This underscores that pretraining effectively encodes a robust behavioral prior that serves as a necessary foundation for downstream adaptation. These patterns closely match the AUC ablations, confirming that our design choices jointly improve both classification performance and the ranking quality measured by KS.

\begin{table*}[htbp]
\centering
\small
\caption{Ablation study (KS) of Q-Anchor Embedding across modality, training method and prompt tuning.}
\label{tab:ablations_ks}
\begin{tabular*}{\textwidth}{@{\extracolsep{\fill}} l *{11}{c} @{}}
\toprule
& \multicolumn{4}{c}{\textbf{User Engagement}} 
& \multicolumn{2}{c}{\textbf{Risk}} 
& \multicolumn{4}{c}{\textbf{Marketing Sensitivity}} 
& \textbf{Avg. KS} \\
\cmidrule(lr){2-5} \cmidrule(lr){6-7} \cmidrule(lr){8-11}

\textbf{Method} 
& \textbf{Active} 
& \textbf{Concert} 
& \textbf{Login} 
& \textbf{Forest} 
& \textbf{Fraud} 
& \textbf{Money} 
& \textbf{TakeOut} 
& \textbf{Brand} 
& \textbf{Promo} 
& \textbf{Value}
& \\
\midrule
\textsc{Q-Anchor} (Base)    
& \underline{0.2228} & \textbf{0.1023} & \textbf{0.5216}	& \underline{0.8437} & \underline{0.7054} & \textbf{0.7450}&  \underline{0.6210} & \textbf{0.4527} & 0.1727 & \underline{0.6567} & \textbf{0.5044}  \\
\midrule
\multicolumn{12}{@{}l@{}}{\textit{\textbf{Ablation on Modality Token}}} \\
w / o User Tok.
& 0.2176	& \underline{0.1021}	& \underline{0.5204}	& \textbf{0.8440}	& \textbf{0.7056}	& 0.7387	& 0.6200	& \underline{0.4309}	& 0.1744	& 0.6478 & 0.5002
\\

w / o Modal Tok.
& 0.2186	& 0.0982	& 0.5192	& 0.8415	& 0.7039	& \underline{0.7407}	& 0.6187	& 0.4321	& \textbf{0.1775}	& \textbf{0.6574} & \underline{0.5008}
\\
w / o Modal \& User Tok.
& 0.2181	& 0.0966	& 0.5169	& 0.8411	& 0.7030	& 0.7402	& 0.6184	& 0.4246	& 0.1654	& 0.6414 & 0.4966  \\

\midrule
\multicolumn{12}{@{}l@{}}{\textit{\textbf{Ablation on Training Method}}} \\
w / o Filter 
& 0.2082	
& 0.0982	
& 0.5176	
& 0.8370	
& 0.6986	
& 0.7386	
& 0.6071	
& 0.3824	
& \underline{0.1755}
& 0.6133
& 0.4877 \\

w / o NTP
& 0.2255	
& 0.0952	
& 0.5150	
& 0.8432 
& 0.7024	
& 0.7404	
& \textbf{0.6225}	
& 0.4298	
& 0.1524	
& 0.6346
& 0.4961
\\

w / o Contrastive
& 0.2028	
& 0.0662	
& 0.5213	
& 0.7977	
& 0.6690	
& 0.6752	
& 0.4534 
& 0.2169	
& 0.1423	
& 0.4702
& 0.4215
\\
% \bottomrule
\midrule
% \midrule
\textsc{Q-Anchor} (Prompt Tuned)      
& \textbf{0.2412} & \textbf{0.1177} & \textbf{0.5225} & \textbf{0.8480} & \textbf{0.7086} & \textbf{0.7573} & \textbf{0.6212} & \textbf{0.5564} & \textbf{0.1962} & \textbf{0.6978} & \textbf{0.5267} \\
\midrule
\multicolumn{12}{@{}l@{}}{\textit{\textbf{Ablation on Prompt Tuning}}} \\
w / o pretrain
& 0.2200
& 0.0931
& 0.4522
& 0.7822
& 0.6953
& 0.5645
& 0.6202
& 0.4436
& 0.1528
& 0.6550
& 0.4679 \\

\bottomrule
\end{tabular*}
\end{table*}

\subsection{Scalability of Q-Anchor Embedding (Base)}
\label{subapp:per_scenario_base}
\paragraph{\textbf{Scalability performance across pretraining data scales.}}
Table~\ref{tab:data_scaling_scalability} evaluates \textit{Q-Anchor (Base)} under increasing pretraining data scales (10k--50k steps, i.e., 20.5M--102.4M samples with batch size 2048). Overall, performance improves steadily as we scale up pretraining, with the best overall results achieved at \textbf{50k steps}: Avg. AUC increases from 0.8029 (10k) to \textbf{0.8105}, and Avg. KS rises from 0.4895 to \textbf{0.5044}. Gains are particularly clear in high-signal domains such as Risk and Marketing (e.g., \textit{Money} AUC: 0.9340 $\rightarrow$ \textbf{0.9382}; \textit{Value} AUC: 0.8869 $\rightarrow$ \textbf{0.9049}), suggesting that larger pretraining exposes the encoder to more diverse long-tail behavioral patterns and improves query-conditioned ranking separability. While a few scenarios saturate early (e.g., \textit{Forest}), the overall trend indicates that scaling pretraining data consistently strengthens both classification (AUC) and ranking quality (KS); therefore, we adopt \textbf{50k} as our default setting in all subsequent experiments.

\begin{table*}[htbp]
\centering
\small
\caption{Scalability performance of Q-Anchor Embedding (Base) across pretraining steps (10k--50k) and corresponding data scales (20.5M--102.4M samples, batch size=2048) (AUC \& KS).}
\label{tab:data_scaling_scalability}
\begin{tabular*}{\textwidth}{@{\extracolsep{\fill}} l *{11}{c} @{}}
\toprule
& \multicolumn{4}{c}{\textbf{User Engagement}} 
& \multicolumn{2}{c}{\textbf{Risk}} 
& \multicolumn{4}{c}{\textbf{Marketing Sensitivity}} 
& \textbf{Avg.} \\
\cmidrule(lr){2-5} \cmidrule(lr){6-7} \cmidrule(lr){8-11}
\textbf{Steps / Data scale} 
& \textbf{Active} 
& \textbf{Concert} 
& \textbf{Login} 
& \textbf{Forest} 
& \textbf{Fraud} 
& \textbf{Money} 
& \textbf{TakeOut} 
& \textbf{Brand} 
& \textbf{Promo} 
& \textbf{Value}
& \\
\midrule

\multicolumn{12}{@{}l@{}}{\textit{\textbf{AUC Performance}}} \\
w/ 10k / 20.48M 
& 0.6443 & 0.5630 & 0.8419 & \textbf{0.9707} & 0.9170 & 0.9340 & 0.8759 & 0.7750 & 0.6206 & 0.8869 & 0.8029 \\
w/ 20k / 40.96M  
& 0.6481 & 0.5713 & \textbf{0.8429} & 0.9705 & 0.9198 & 0.9362 & 0.8776 & \underline{0.7988} & \textbf{0.6259} & 0.8999 & 0.8091 \\
w/ 30k / 61.44M  
& 0.6472 & \underline{0.5735} & \textbf{0.8429} & \underline{0.9706} & 0.9203 & 0.9365 & 0.8782 & \textbf{0.8027} & \underline{0.6229} & 0.9041 & \underline{0.8099} \\
w/ 40k / 81.92M  
& \underline{0.6532} & \textbf{0.5742} & 0.8419 & 0.9704 & \underline{0.9204} & \underline{0.9369} & \underline{0.8792} & 0.7986 & 0.6159 & \underline{0.9045} & 0.8095 \\
w/ 50k / 102.4M  
& \textbf{0.6571} & 0.5739 & \underline{0.8420} & 0.9700 & \textbf{0.9218} & \textbf{0.9382} & \textbf{0.8799} & 0.7979 & 0.6190 & \textbf{0.9049} & \textbf{0.8105} \\

\midrule
\multicolumn{12}{@{}l@{}}{\textit{\textbf{KS Performance}}} \\
w/ 10k / 20.48M 
& 0.2036 & 0.0857 & 0.5166 & \underline{0.8447} & 0.6950 & 0.7337 & 0.6140 & 0.4099 & 0.1761 & 0.6153 & 0.4895 \\
w/ 20k / 40.96M 
& 0.2110 & 0.0988 & 0.5205 & 0.8445 & 0.7012 & 0.7369 & 0.6167 & 0.4534 & \textbf{0.1848} & 0.6469 & 0.5015 \\
w/ 30k / 61.44M  
& 0.2099 & 0.1008 & \underline{0.5211} & \textbf{0.8459} & \underline{0.7030} & 0.7391 & 0.6173 & \textbf{0.4606} & \underline{0.1789} & 0.6554 & \underline{0.5032} \\
w/ 40k / 81.92M 
& \underline{0.2183} & \underline{0.1015} & 0.5204 & \underline{0.8447} & 0.7025 & \underline{0.7405} & \underline{0.6201} & \underline{0.4536} & 0.1685 & \textbf{0.6572} & 0.5027 \\
w/ 50k / 102.4M 
& \textbf{0.2228} & \textbf{0.1023} & \textbf{0.5216} & 0.8437 & \textbf{0.7054} & \textbf{0.7450} & \textbf{0.6210} & 0.4527 & 0.1727 & \underline{0.6567} & \textbf{0.5044} \\
\bottomrule
\end{tabular*}
\end{table*}

\begin{figure}[htbp]
    \centering
    \begin{subfigure}{0.49\columnwidth} 
        \centering
        \includegraphics[width=\linewidth]{assets/scalability_model.png}
        \caption{Performance (Avg. AUC/KS) vs. model size (0.5B--3B).}
        \label{fig:model_scale}
    \end{subfigure}
    \hfill
    \begin{subfigure}{0.49\columnwidth}
        \centering
        \includegraphics[width=0.88\linewidth]{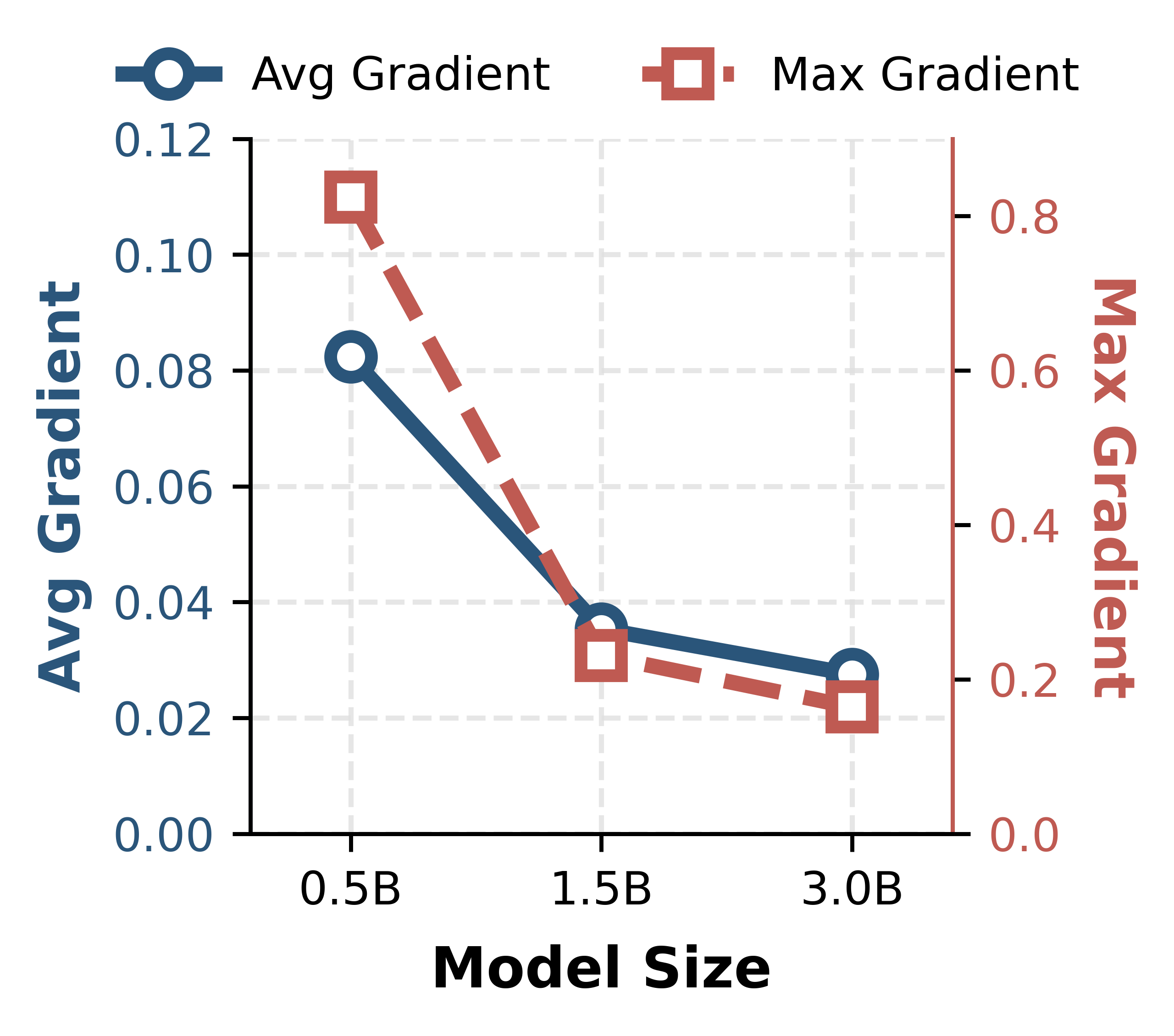}
        \caption{Gradient magnitude decreases with model scale.}
        \label{fig:gradient_scale}
    \end{subfigure}
    \caption{Scalability analysis of \textsc{Q-Anchor (Base)}. (a) Smaller 0.5B backbone achieves best Avg. AUC/KS; larger models show no consistent gains. (b) Gradient magnitude diminishes as model size increases, explaining the non-monotonic performance trend.}
    \label{fig:gradient_scale}
\end{figure}

\paragraph{\textbf{Scalability performance across model scales.}}
Table~\ref{tab:model_scaling_base} compares \textit{Q-Anchor (Base)} under different backbone scales (0.5B--3B). We observe that scaling up the model does not monotonically improve performance: the \textbf{0.5B} backbone achieves the best overall results (Avg. AUC/KS: \textbf{0.8105}/\textbf{0.5044}), while larger models (1.5B and 3B) provide no consistent gains and even slightly regress on several scenarios (e.g., \textit{Brand} and \textit{Promo}). This non-monotonic scaling behavior is aligned with prior findings \cite{jiang2024scaling} that, under fixed data and training budgets, larger encoders may not translate into better embedding quality and can be harder to optimize for sentence-level embedding objectives~\cite{jiang2024scaling}. From an industrial perspective, the \textbf{0.5B} configuration is also preferable for deployment due to its substantially lower latency and serving cost, making it a better accuracy--efficiency trade-off for high-throughput online Alipay financial and risk systems. 

To investigate the non-monotonic performance observed across different model scales, we analyze the gradient dynamics during the late stages of training. As shown in Figure~\ref{fig:gradient_scale}, both the average and maximum gradient magnitudes decrease sharply as the model size increases: the average gradient drops from 0.082 (0.5B) to 0.035 (1.5B) and further to 0.028 (3B), while the maximum gradient similarly decreases from 0.824 to 0.231 and 0.164, respectively. This empirical evidence indicates that under fixed data budgets, larger encoders produce increasingly flat optimization landscapes for sentence-level embedding objectives, leading to a “saturation” of learning signals. Consistent with prior findings \cite{jiang2024scaling}, this explains why larger models, despite their generative capacity, are harder to optimize for discriminative embedding alignment. From an industrial perspective, the 0.5B configuration achieves the best accuracy–efficiency trade-off, offering stronger gradients, better embedding quality, and significantly lower serving latency and cost, making it ideal for high-throughput deployment in Alipay's production systems.

\begin{table*}[htbp]
\centering
\small
\caption{Scalability performance of Q-Anchor Embedding (Base) across model scales (0.5B–3B).}
\label{tab:model_scaling_base}
\begin{tabular*}{\textwidth}{@{\extracolsep{\fill}} l *{11}{c} @{}}
\toprule
& \multicolumn{4}{c}{\textbf{User Engagement}} 
& \multicolumn{2}{c}{\textbf{Risk}} 
& \multicolumn{4}{c}{\textbf{Marketing Sensitivity}} 
& \textbf{Avg.} \\
\cmidrule(lr){2-5} \cmidrule(lr){6-7} \cmidrule(lr){8-11}
\textbf{Model} 
& \textbf{Active} 
& \textbf{Concert} 
& \textbf{Login} 
& \textbf{Forest} 
& \textbf{Fraud} 
& \textbf{Money} 
& \textbf{TakeOut} 
& \textbf{Brand} 
& \textbf{Promo} 
& \textbf{Value}
& \\
\midrule

\multicolumn{12}{@{}l@{}}{\textit{\textbf{AUC Performance}}} \\
\textsc{Q-Anchor}-0.5B (Base) & \textbf{0.6571} & \textbf{0.5739} & 0.8420 & 0.9700 & \underline{0.9218} & \textbf{0.9382} & \textbf{0.8799} & \textbf{0.7979} & \textbf{0.6190} & \textbf{0.9049} & \textbf{0.8105} \\
\textsc{Q-Anchor}-1.5B (Base) & \underline{0.6436} & \underline{0.5693} & \underline{0.8444} & \underline{0.9706} & \textbf{0.9220} & \underline{0.9376} & 0.8759 & \underline{0.7864} & \underline{0.6156} & \underline{0.9000} & \underline{0.8065} \\
\textsc{Q-Anchor}-3B (Base)   & 0.6404 & 0.5626 & \textbf{0.8449} & \textbf{0.9712} & \textbf{0.9220} & 0.9370 & \underline{0.8777} & 0.7824 & 0.6066 & 0.8989 & 0.8044 \\

\midrule
\multicolumn{12}{@{}l@{}}{\textit{\textbf{KS Performance}}} \\
\textsc{Q-Anchor}-0.5B (Base) & \textbf{0.2228} & \textbf{0.1023} & 0.5216 & 0.8437 & \textbf{0.7054} & \textbf{0.7450} & \textbf{0.6210} & \textbf{0.4527} & \textbf{0.1727} & \textbf{0.6567} & \textbf{0.5044} \\
\textsc{Q-Anchor}-1.5B (Base) & \underline{0.2010} & \underline{0.0969} & \underline{0.5223} & \underline{0.8442} & \underline{0.7052} & \underline{0.7428} & 0.6151 & \underline{0.4295} & \underline{0.1657} & \underline{0.6486} & \underline{0.4971} \\
\textsc{Q-Anchor}-3B (Base)   & 0.1957 & 0.0875 & \textbf{0.5234} & \textbf{0.8457} & 0.7045 & 0.7407 & \underline{0.6171} & 0.4231 & 0.1493 & 0.6454 & 0.4932 \\
\bottomrule
\end{tabular*}
\end{table*}

% \newpage
\subsection{Scalability performance of Q-Anchor Embedding (Prompt Tuned)}
\label{subapp:per_scenario_pt}
\paragraph{\textbf{Scalability performance across learnable token scales.} }
Table~\ref{tab:t_tokens_scalability} studies the scalability of prompt tuning by varying the number of learnable prompt tokens from 1 to 16. Overall, performance improves rapidly when increasing tokens from 1 to 6, and then largely saturates: the best average performance is achieved at \textbf{6 prompt tokens} (Avg. AUC/KS: \textbf{0.8225}/\textbf{0.5267}). This suggests that a small number of learnable tokens is sufficient to provide scenario-specific conditioning, while additional tokens bring diminishing returns and may introduce optimization noise on some tasks. Given this clear accuracy--efficiency trade-off, we adopt \textbf{6 prompt tokens} as the default configuration in all subsequent experiments.
\begin{table*}[htbp]
\centering
\caption{Scalability performance of Q-Anchor Embedding (Prompt Tuned) across learnable token scales (1–16) for prompt tuning (Ours: 6 prompt tokens).}
\label{tab:t_tokens_scalability}
\begin{tabular*}{\textwidth}{@{\extracolsep{\fill}} l *{11}{c} @{}}
\toprule
& \multicolumn{4}{c}{\textbf{User Engagement}} 
& \multicolumn{2}{c}{\textbf{Risk}} 
& \multicolumn{4}{c}{\textbf{Marketing Sensitivity}} 
& \textbf{Avg.} \\
\cmidrule(lr){2-5} \cmidrule(lr){6-7} \cmidrule(lr){8-11}
\textbf{Token} 
& \textbf{Active} 
& \textbf{Concert} 
& \textbf{Login} 
& \textbf{Forest} 
& \textbf{Fraud} 
& \textbf{Money} 
& \textbf{TakeOut} 
& \textbf{Brand} 
& \textbf{Promo} 
& \textbf{Value}
& \\
\midrule

\multicolumn{12}{@{}l@{}}{\textit{\textbf{AUC Performance}}} \\
w / 1 prompt tok.  
& 0.6513 & 0.5746 & 0.8408 & 0.9702 & 0.9215 & 0.9424 & 0.8798 & 0.8290 & 0.6250 & 0.9114 & 0.8146 \\
w / 2 prompt tok.   
& 0.6576 & 0.5755 & 0.8408 & 0.9696 & 0.9221 & 0.9407 & \underline{0.8810} & 0.8380 & 0.6269 & 0.9170 & 0.8169 \\
w / 4 prompt tok.   
& \underline{0.6625} & \underline{0.5792} & \textbf{0.8450} & 0.9715 & 0.9220 & \underline{0.9430} & 0.8804 & 0.8501 & 0.6321 & 0.9155 & 0.8201 \\
w / 6 prompt tok.   
& \textbf{0.6678} & \textbf{0.5844} & \underline{0.8443} & \underline{0.9716} & \textbf{0.9242} & \textbf{0.9439} & \textbf{0.8811} & 0.8535 & \underline{0.6350} & \underline{0.9194} & \textbf{0.8225} \\
w / 8 prompt tok.   
& 0.6612 & 0.5774 & 0.8432 & 0.9704 & 0.9228 & 0.9409 & 0.8801 & \underline{0.8575} & 0.6310 & \textbf{0.9205} & 0.8205 \\
w / 16 prompt tok.  
& 0.6649 & 0.5768 & 0.8435 & \textbf{0.9740} & \underline{0.9230} & 0.9389 & 0.8793 & \textbf{0.8585} & \textbf{0.6351} & \textbf{0.9205} & \underline{0.8215} \\

\midrule
\multicolumn{12}{@{}l@{}}{\textit{\textbf{KS Performance}}} \\
w / 1 prompt tok.   
& 0.2176 & 0.1031 & 0.5208 & 0.8444 & 0.7053 & 0.7576 & \underline{0.6228} & 0.5152 & 0.1790 & 0.6742 & 0.5140 \\
w / 2 prompt tok.   
& 0.2240 & 0.1032 & 0.5202 & 0.8433 & 0.7054 & 0.7529 & \textbf{0.6233} & 0.5292 & 0.1865 & 0.6893 & 0.5177 \\
w / 4 prompt tok.    
& 0.2323 & 0.1068 & 0.5222 & \underline{0.8509} & 0.7056 & \textbf{0.7593} & 0.6208 & 0.5537 & 0.1928 & 0.6833 & 0.5228 \\
w / 6 prompt tok.    
& \textbf{0.2412} & \textbf{0.1177} & \underline{0.5225} & 0.8480 & \textbf{0.7086} & \underline{0.7573} & 0.6212 & 0.5564 & \underline{0.1962} & 0.6978 & \textbf{0.5267} \\
w / 8 prompt tok.   
& 0.2317 & \underline{0.1078} & \textbf{0.5229} & 0.8432 & 0.7073 & 0.7520 & 0.6219 & \underline{0.5646} & 0.1904 & \textbf{0.6995} & 0.5241 \\
w / 16 prompt tok.   
& \underline{0.2351} & 0.1058 & \textbf{0.5229} & \textbf{0.8577} & \underline{0.7082} & 0.7461 & 0.6209 & \textbf{0.5694} & \textbf{0.1975} & \underline{0.6988} & \underline{0.5262} \\
\bottomrule
\end{tabular*}
\end{table*}

\paragraph{\textbf{Scalability performance across learning step scales.}}
Table~\ref{tab:pt_steps} reports the scalability of prompt-tuned Q-Anchor embedding under different training-step budgets (100--500, step size = 100). As the number of steps increases, performance improves consistently on both AUC and KS, with the average AUC rising from 0.8159 (100 steps) to 0.8225 (500 steps) and the average KS from 0.5141 to 0.5267. Notably, 500 steps achieves the best overall results, delivering the top average AUC/KS and leading most tasks, while 400 steps is typically the strongest runner-up. These results indicate that our method benefits from longer optimization yet remains robust under smaller step budgets, showing stable scalability across training steps.

\begin{table*}[htbp]
\centering
\caption{Scalability of Q-Anchor Embedding (Prompt Tuned) across training steps (100–500) for prompt tuning (Ours: 500 steps).}
\label{tab:pt_steps}
\begin{tabular*}{\textwidth}{@{\extracolsep{\fill}} l *{10}{c} c @{}}
\toprule
& \multicolumn{4}{c}{\textbf{User Engagement}} 
& \multicolumn{2}{c}{\textbf{Risk}} 
& \multicolumn{4}{c}{\textbf{Marketing Sensitivity}} 
& \textbf{Avg.} \\
\cmidrule(lr){2-5} \cmidrule(lr){6-7} \cmidrule(lr){8-11}
\textbf{Steps}
& \textbf{Active} 
& \textbf{Concert} 
& \textbf{Login} 
& \textbf{Forest} 
& \textbf{Fraud} 
& \textbf{Money} 
& \textbf{TakeOut} 
& \textbf{Brand} 
& \textbf{Promo} 
& \textbf{Value}
& \\
\midrule

\multicolumn{12}{@{}l@{}}{\textit{\textbf{AUC Performance}}} \\
w / 100   & 0.6541 & 0.5775 & 0.8441 & 0.9703 & 0.9230 & 0.9382 & 0.8787 & 0.8329 & 0.6308 & 0.9091 & 0.8159 \\
w / 200   & 0.6601 & 0.5774 & 0.8432 & 0.9701 & \underline{0.9237} & 0.9403 & 0.8796 & 0.8454 & 0.6303 & 0.9139 & 0.8184 \\
w / 300   & 0.6622 & 0.5806 & \underline{0.8434} & 0.9698 & 0.9234 & 0.9428 & \underline{0.8810} & 0.8484 & 0.6281 & \underline{0.9167} & 0.8196 \\
w / 400   & \underline{0.6657} & \underline{0.5821} & \underline{0.8434} & \textbf{0.9718} & 0.9236 & \underline{0.9433} & 0.8799 & \underline{0.8513} & \underline{0.6348} & \underline{0.9167} & \underline{0.8213} \\
w / 500 & \textbf{0.6678} & \textbf{0.5844} & \textbf{0.8443} & \underline{0.9716} & \textbf{0.9242} & \textbf{0.9439} & \textbf{0.8811} & \textbf{0.8535} & \textbf{0.6350} & \textbf{0.9194} & \textbf{0.8225} \\
\midrule

\multicolumn{12}{@{}l@{}}{\textit{\textbf{KS Performance}}} \\
w / 100   & 0.2212 & 0.1059 & 0.5225 & 0.8429 & 0.7064 & 0.7425 & 0.6191 & 0.5195 & 0.1913 & 0.6701 & 0.5141 \\
w / 200  & 0.2287 & 0.1041 & 0.5204 & 0.8433 & 0.7076 & 0.7488 & 0.6210 & 0.5452 & 0.1912 & 0.6797 & 0.5190 \\
w / 300   & 0.2317 & 0.1100 & 0.5227 & 0.8418 & 0.7076 & 0.7543 & 0.6227 & 0.5478 & 0.1864 & 0.6866 & 0.5212 \\
w / 400  & \underline{0.2364} & \underline{0.1116} & \textbf{0.5227} & \underline{0.8466} & \underline{0.7079} & \underline{0.7553} & \textbf{0.6214} & \underline{0.5538} & \underline{0.1958} & \underline{0.6917} & \underline{0.5243} \\
w / 500 & \textbf{0.2412} & \textbf{0.1177} & \underline{0.5225} & \textbf{0.8480} & \textbf{0.7086} & \textbf{0.7573} & \underline{0.6212} & \textbf{0.5564} & \textbf{0.1962} & \textbf{0.6978} & \textbf{0.5267} \\
\bottomrule
\end{tabular*}
\end{table*}

\subsection{PCA Visualization of Universal and Prompt-tuned Representation}
\label{subapp:PCA}
\begin{figure}[!t]
\captionsetup[subfigure]{skip=1pt}
    \centering
    \begin{subfigure}[b]{\linewidth}
        \centering
        \includegraphics[width=\linewidth]{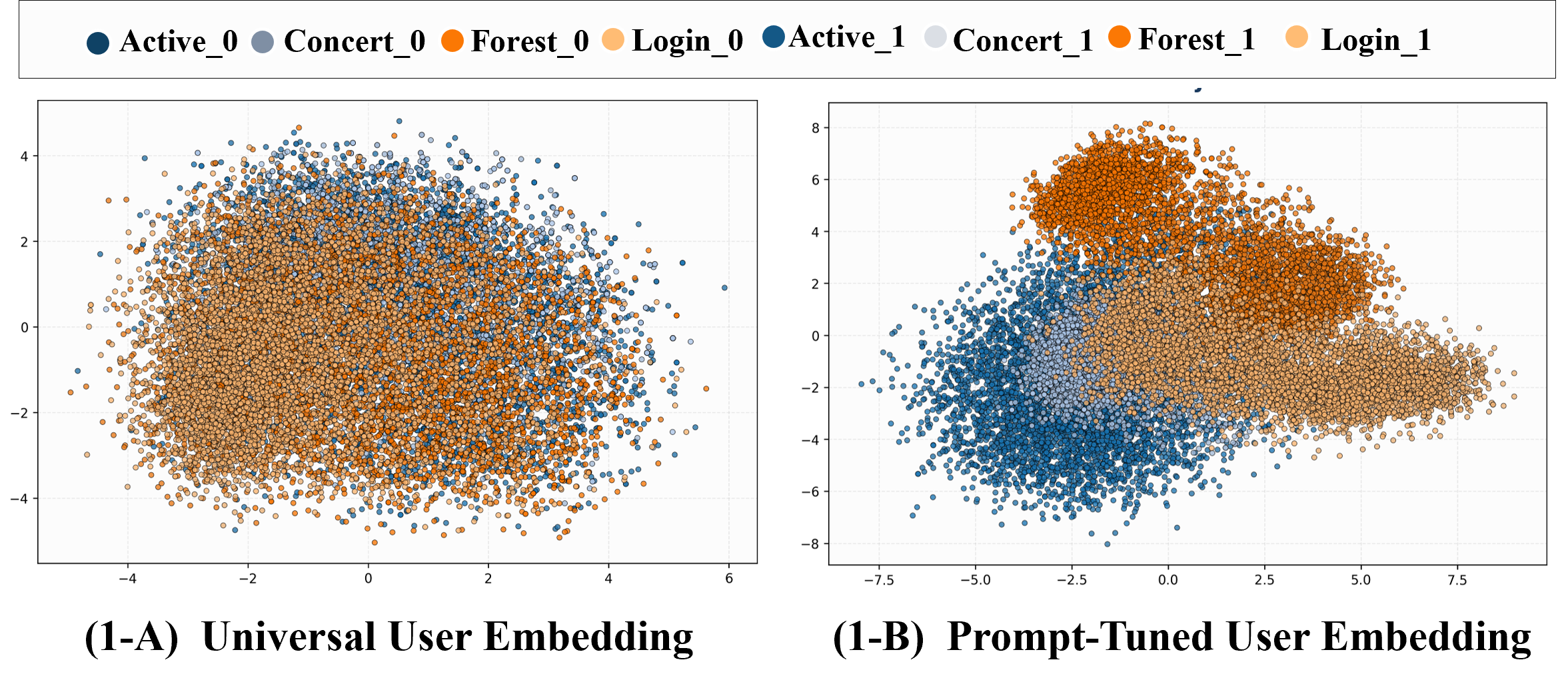}
        \caption{User Engagement}
        \label{fig:1}
    \end{subfigure}
    % \hfill    
    \begin{subfigure}[b]{\linewidth}
        \centering
        \includegraphics[width=\linewidth]{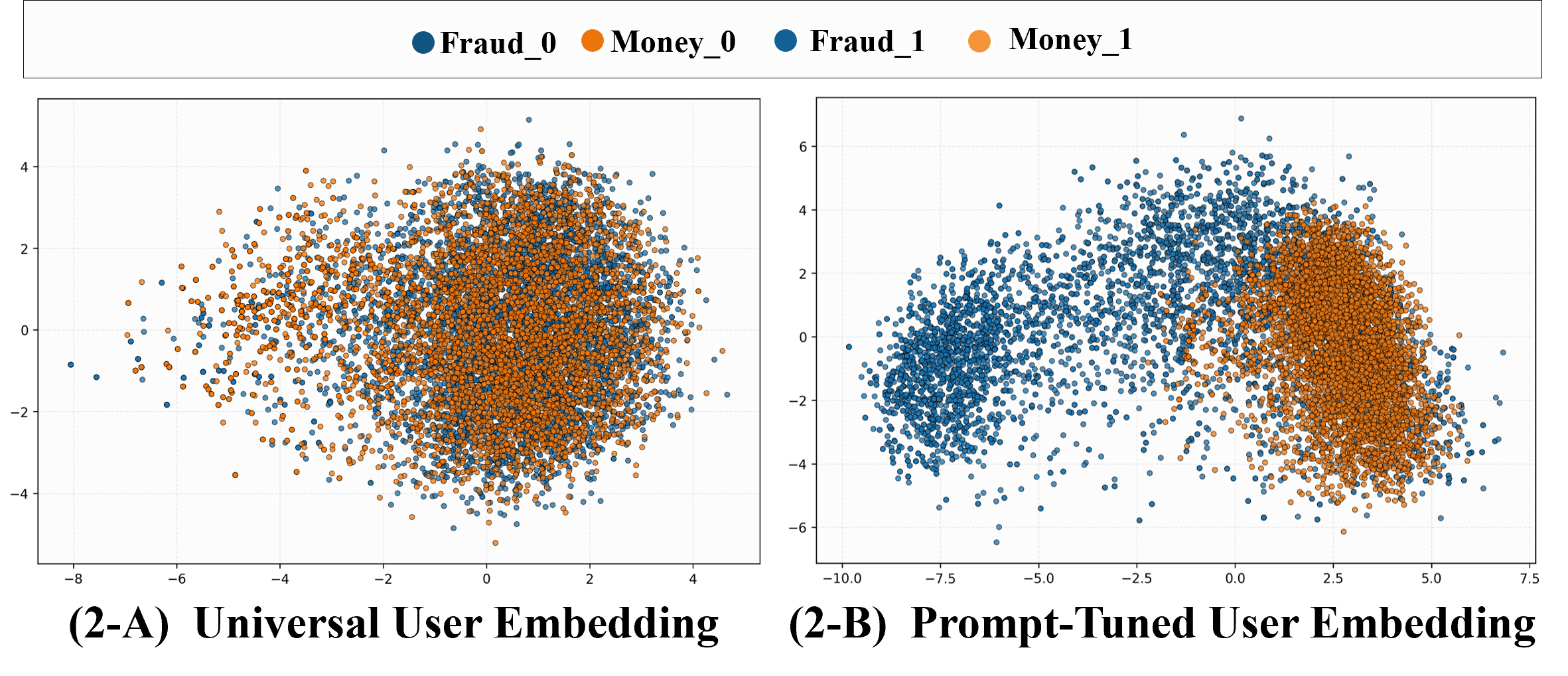}
        \caption{Risk}
        \label{fig:2}
    \end{subfigure}
    
    % \hfill    
    
    \begin{subfigure}[b]{\linewidth}
        \centering
        \includegraphics[width=\linewidth]{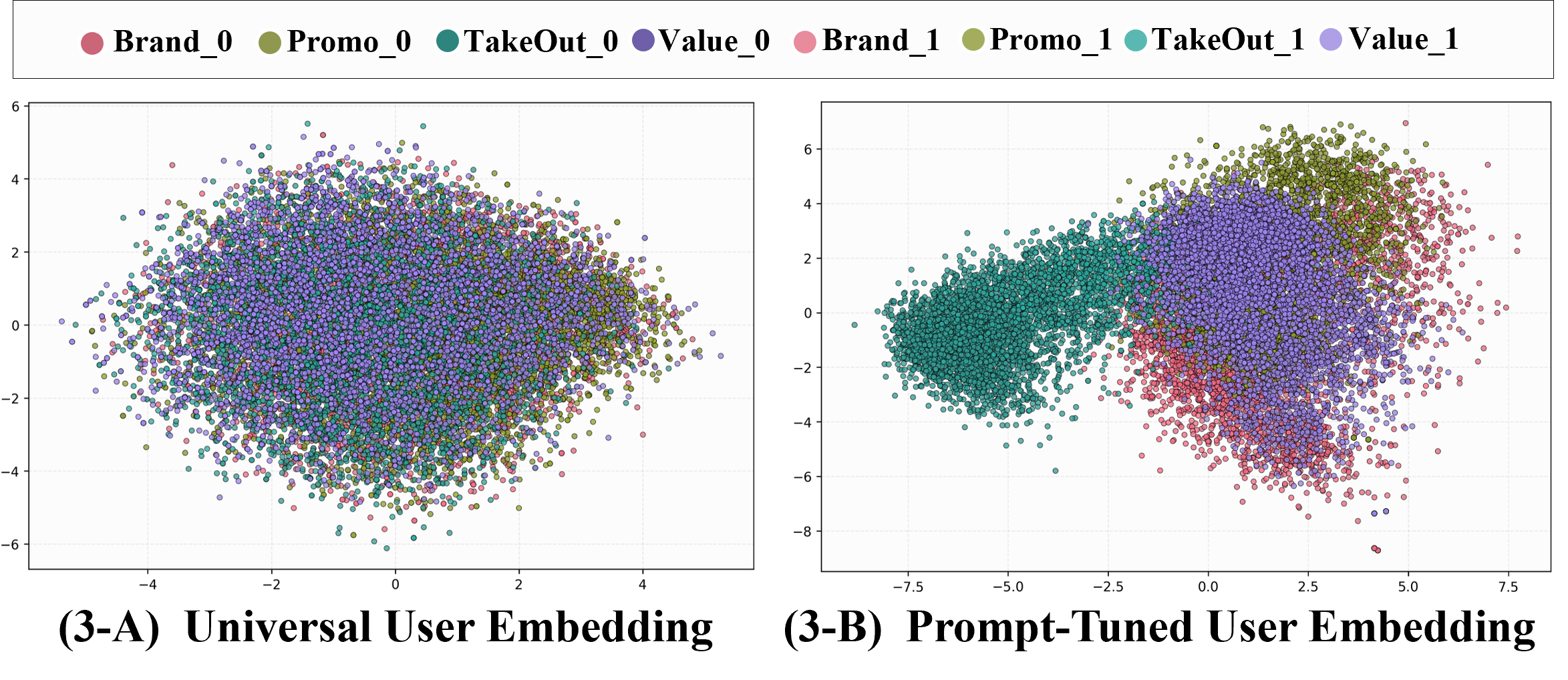}
        \caption{Marketing Sensitivity}
        \label{fig:3}
    \end{subfigure}
    
    \caption{PCA Visualization of universal and prompt-tuned representation of 10 scenarios.}
    \label{fig:pca}
\end{figure}

To complement our t-SNE analysis and strengthen the reliability of qualitative insights, we further visualize the universal and prompt-tuned representations using PCA (Fig.~\ref{fig:pca}). Across all three domains—User Engagement, Risk, and Marketing Sensitivity—the prompt-tuned embeddings consistently exhibit tighter intra-class clustering and clearer separation between positive and negative samples compared to the base universal representation. This effect is especially pronounced in high-stakes scenarios such as Brand and Money, where subtle behavioral signals determine model performance. The convergence of t-SNE and PCA visualizations with consistent quantitative improvements in AUC and KS underscores that lightweight scenario-specific conditioning effectively reshapes the embedding geometry to better align with downstream decision boundaries—without modifying the underlying architecture.

\begin{table}[H]
\centering
\caption{Offline performance of \textsc{Q-Anchor} embeddings. Base uses universal embeddings; Prompt Tuned adds lightweight, scenario-specific soft prompts. Both scenarios report AUC and KS, with Delinquency primarily optimizing KS.}
\label{tab:offline_results}
\small
\begin{tabular}{lcc}
\toprule
\textbf{Method} & \textbf{AUC} & \textbf{KS} \\
\midrule
\multicolumn{3}{l}{\textbf{Scenario: IVR Response}} \\
Business SOTA & 0.6362 & 0.2411 \\
\textbf{\textsc{Q-Anchor} (Base)} & \underline{0.6534} & \underline{0.2682} \\
\textbf{\textsc{Q-Anchor} (Prompt Tuned)} & \textbf{0.6863} & \textbf{0.3016} \\
\midrule
\multicolumn{3}{l}{\textbf{Scenario: Delinquency}} \\
Business SOTA & - & 0.1499 \\
\textbf{\textsc{Q-Anchor} (Base)} & - & \underline{0.1534} \\
\textbf{\textsc{Q-Anchor} (Prompt Tuned)} & - & \textbf{0.1700} \\
\bottomrule
\end{tabular}
\end{table}

\section{Pre-Deployment Offline Performance on Downstream Business}
\label{app:offline}
Before online deployment, we evaluate \textsc{Q-Anchor} embeddings against current production SOTA (highly optimized, domain-specific handcrafted features and specialized models) across two disparate tasks: \textit{IVR Response Prediction} and \textit{Credit Delinquency Identification}. Table~\ref{tab:offline_results} summarizes the results.

\textbf{Robustness of Universal Representations.} 
Even in its \textit{Base} form, \textsc{Q-Anchor} consistently outperforms the Business SOTA across all metrics. Specifically, in the IVR scenario, it achieves a \textbf{2.71\% absolute lift} in KS over handcrafted features. This confirms that our pre-training objective effectively distills a foundational understanding of user behavior that generalizes across different financial contexts without any scenario-specific adaptation.

\textbf{Steering Performance with Minimal Overhead.} 
By introducing scenario-specific soft prompts, \textbf{\textsc{Q-Anchor} (Prompt Tuned)} further unlocks the latent potential of the universal backbone. It reaches a \textbf{0.3016 KS (+6.05\% over SOTA)} in IVR and a \textbf{0.1700 KS (+2.01\% over SOTA)} in delinquency prediction. 

Crucially, this substantial performance gain is achieved with \textbf{negligible incremental cost}. As \textsc{Q-Anchor} reuses the expensive user-prefix KV-cache, the prompt tuning only involves optimizing a few learnable vectors at the query suffix. This architecture allows us to "steer" the same multi-billion parameter representation toward diverse business goals, achieving a superior balance between high-precision task adaptation and industrial-scale computational efficiency.

\section{Deployment Detail of Q-Anchor Embedding}
\label{app:deploy}
To serve embedding requests at Alipay scale, we deploy \textsc{Q-Anchor Embedding} with an \emph{incremental update} pipeline that decouples heavy user-history encoding from lightweight scenario querying.
Each day, the system ingests newly arrived multi-modal behaviors and updates the user profile by \emph{re-encoding only the affected modalities} (i.e., modalities that receive new events) and refreshing their corresponding modality- and period-level summary tokens, rather than reprocessing all modalities and all raw events from the past 90 days end-to-end. For each modality $m\!\in\!\mathcal{M}$, a frozen backbone with a modality-specific event adapter encodes the day's events into event-level tokens, which are further aggregated into a modality summary token $\mathbf{z}^{(\mathrm{mdl})}_m$ (Eq.~\ref{eq:tokens}).
Newly generated tokens are updated into the historical token buffer to form the updated hierarchical prefix $\mathbf{e}_i$, while tokens outside the rolling window are expired. This design keeps representations fresh, bounded, and cost-stable.

For online inference, we exploit \textbf{Query-as-Anchor} with \textbf{prefix KV-cache sharing} (Fig.~\ref{fig:deploy}).
For each user, the prefix $\mathbf{e}_i$ is encoded once to build a shared KV cache that is reused across downstream scenarios.
Given scenario queries $\{q_1,\dots,q_n\}$ (where $n$ is the number of scenarios), we process the queries \emph{sequentially} while reusing the same shared prefix cache.
For each query $q_j$, the model only computes the short query suffix on top of the cached prefix, reducing the marginal per-scenario cost to $O(L_{q_j})$ (and $O(\sum_{j=1}^n L_{q_j})$ for all scenarios) and enabling efficient multi-scenario re-anchoring (e.g., risk, marketing, engagement).
Overall, the deployment provides (i) \textbf{efficient daily refresh} via delta updates and (ii) \textbf{high-throughput multi-scenario serving} by amortizing prefix encoding with KV-cache reuse.

\section{Future Work}

While standard LLM scaling laws \cite{zhangscaling} suggest performance improves with model size, our experiments reveal a Scaling Paradox in user embeddings: larger backbones (1.5B–3B) exhibit gradient attenuation and performance stagnation compared to the 0.5B model, consistent with prior observations \cite{jiang2024scaling}. This indicates that embedding quality may follow a discriminative scaling trajectory, governed by the signal-to-parameter ratio rather than raw parameter count. Future work will investigate gradient recovery and adaptive parameter tuning to overcome optimization plateaus and systematically extend scaling benefits to larger models. Within the standardized, fair comparison setup of this study—without additional hyperparameter or learning-rate tricks—\textsc{Q-Anchor} achieves optimal performance at minimal cost on the 0.5B backbone. Accordingly, this base model will remain the primary configuration for our deployment and large-scale adoption across diverse Alipay scenarios.

\end{document}